\documentclass[10pt,journal,compsoc]{IEEEtran}

\usepackage[nocompress]{cite}

\usepackage{url}
\usepackage{ragged2e}
\usepackage{epsfig}
\usepackage{graphicx}
\usepackage{amsmath,amssymb} 
\usepackage{mathptmx}      
\usepackage{subfigure}
\usepackage{algorithm}
\usepackage{algorithmicx}
\usepackage{algpseudocode}
\usepackage{graphics}
\usepackage{mathrsfs}
\usepackage{threeparttable}
\usepackage{color}
\usepackage[normalem]{ulem}
\usepackage{multirow}
\usepackage{float}
\usepackage{amsfonts}
\usepackage{bm}
\usepackage{array}
\usepackage[table]{xcolor}
\usepackage{colortbl}
\usepackage{pifont}
\usepackage{diagbox}
\usepackage{rotating}
\usepackage{booktabs}
\usepackage{overpic}
\usepackage{textcomp}
\usepackage{contour}
\usepackage{enumitem}
\usepackage{stfloats}
\usepackage[colorlinks=true,breaklinks=true,urlcolor=magenta]{hyperref}

\def\ie{\emph{i.e.}}
\def\eg{\emph{e.g.}}
\def\etc{\emph{etc}}
\def\etal{{\em et al.~}}

\newcommand{\sArt}{state-of-the-art }
\newcommand{\myPara}[1]{\vspace{6pt}\noindent\textbf{#1}\qquad }

\definecolor{bblue}{rgb}{0,150,230}
\definecolor{mygray}{gray}{.92}

\newcommand{\figref}[1]{Fig.~\ref{#1}}
\newcommand{\tabref}[1]{Table~\ref{#1}}

\newcommand{\secref}[1]{Section \ref{#1}}

\newcommand{\tabincell}[2]{\begin{tabular}{@{}#1@{}}#2\end{tabular}}

\newcommand{\supp}[1]{#1}

\newcommand{\Rev}[1]{\textcolor{black}{#1}}

\def\ourdataset{\textit{COD10K}}
\def\ournewmodel{\emph{SINet}}
\def\baselineN{12}

\graphicspath{{./Imgs/}}
\DeclareGraphicsExtensions{.jpg,.pdf,.png}

\RequirePackage{silence}
\begin{document}

\title{Concealed Object Detection}

\author{Deng-Ping~Fan,~
        Ge-Peng Ji,~
        Ming-Ming Cheng,~
        and Ling Shao
\IEEEcompsocitemizethanks{
\IEEEcompsocthanksitem Deng-Ping Fan is with the CS, Nankai University,
Tianjin, China, and also with the Inception Institute of Artificial 
Intelligence, Abu Dhabi, UAE. (E-mail: dengpfan@gmail.com)
\IEEEcompsocthanksitem Ge-Peng Ji and Ling Shao are with the 
Inception Institute of Artificial Intelligence, Abu Dhabi, UAE. 
(E-mail: gepeng.ji@inceptioniai.org; ling.shao@inceptioniai.org)
\IEEEcompsocthanksitem Ming-Ming Cheng is with the CS, Nankai University,
Tianjin, China. (E-mail: cmm@nankai.edu.cn)
\IEEEcompsocthanksitem A preliminary version of this work has appeared in 
CVPR 2020~\cite{fan2020camouflaged}.
\IEEEcompsocthanksitem The major part of this work was done 
in Nankai University.
\IEEEcompsocthanksitem Ming-Ming Cheng is the corresponding author. 
}
}

\markboth{IEEE TRANSACTIONS ON PATTERN ANALYSIS AND MACHINE INTELLIGENCE}%
{Fan \MakeLowercase{\textit{et al.}}: Concealed Object Detection}

\IEEEtitleabstractindextext{%
\begin{abstract} \justifying
We present the first systematic study on concealed object detection (COD), 
which aims to identify objects that are \Rev{visually embedded} 
in their background.
The high intrinsic similarities between the concealed objects and 
their background make COD far more challenging 
than traditional object detection/segmentation.
To better understand this task, we collect a large-scale dataset, 
called \textbf{COD10K}, 
which consists of 10,000 images covering concealed objects 
in diverse real-world scenarios from 78 object categories. 
Further, we provide rich annotations including object categories, 
object boundaries, challenging attributes, object-level labels, 
and instance-level annotations. 
Our \ourdataset~is the largest COD dataset to date, 
with the richest annotations, 
which enables comprehensive concealed object understanding and 
can even be used to help progress several other vision tasks, 
such as detection, segmentation, classification \etc.  
Motivated by how animals hunt in the wild, 
we also design a simple but strong baseline for COD, 
termed the Search Identification Network (\textbf{\ournewmodel}).
Without any bells and whistles, \ournewmodel~outperforms \Rev{twelve} cutting-edge baselines on all datasets tested,
making them robust, general architectures that could serve as 
catalysts for future research in COD. 
Finally, we provide some interesting findings, 
and highlight several potential applications and future directions.
To spark research in this new field, our code, dataset, 
and online demo are available at our project page: \url{http://mmcheng.net/cod}.
\end{abstract}

}

\maketitle

\IEEEdisplaynontitleabstractindextext

\IEEEpeerreviewmaketitle

\IEEEraisesectionheading{\section{Introduction}\label{sec:introduction}}

\begin{figure}[t!]
  \centering
  \includegraphics[width=\columnwidth]{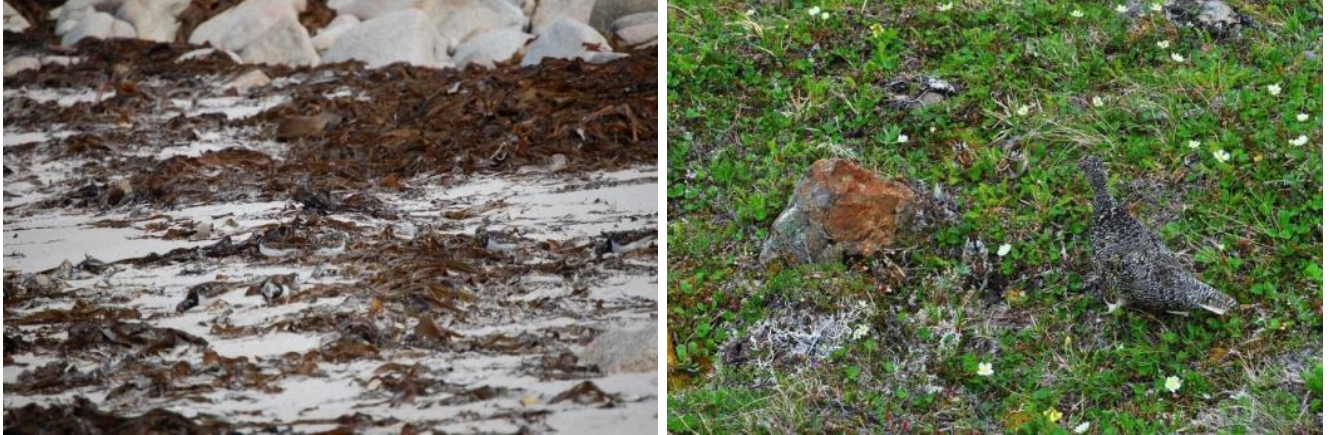}\\
  \vspace{-8pt}
  \caption{\textbf{Examples of background matching camouflage (BMC).} 
    There are seven and six birds for the left and right image, respectively. 
    Answers in color are shown in \figref{fig:Answer}.
  }\label{fig:COD10KExample}
\end{figure}

\IEEEPARstart{C}{an} you find the concealed object(s) in each image of 
\figref{fig:COD10KExample} within 10 seconds? 
Biologists refer to this as \emph{background matching camouflage (BMC)}
\cite{cuthill2005disruptive}, 
where one or more objects attempt to adapt their coloring to match 
``seamlessly'' with the surroundings in order 
to avoid detection~\cite{owens2014camouflaging}.
Sensory ecologists~\cite{stevens2008animal} have found that this BMC strategy 
works by deceiving the visual perceptual system of the observer.
Naturally, addressing \emph{concealed object detection} 
(\textbf{COD}\footnote{We define COD as segmenting objects or stuff 
(amorphous regions~\cite{kirillov2019panoptic}) 
that have a similar pattern, \eg, \emph{texture}, \emph{color}, 
\emph{direction}, \etc., to their natural or man-made environment. 
In the rest of the paper, for convenience, the concealed object segmentation 
is considered identical to COD and used interchangeably.}) 
requires a significant amount of visual perception
\cite{troscianko2008camouflage} knowledge.
\Rev{Understanding COD has not only scientific value in itself,} but it also 
important for applications in many fundamental fields, 
such as computer vision 
(\eg, for search-and-rescue work, or rare species discovery), 
medicine (\eg, polyp segmentation~\cite{fan2020pranet}, 
lung infection segmentation~\cite{fan2020inf}), 
agriculture (\eg, locust detection to prevent invasion), 
and art (\eg, recreational art~\cite{chu2010camouflage}).

\begin{figure}[t!]
  \centering
  \begin{overpic}[width=\columnwidth]{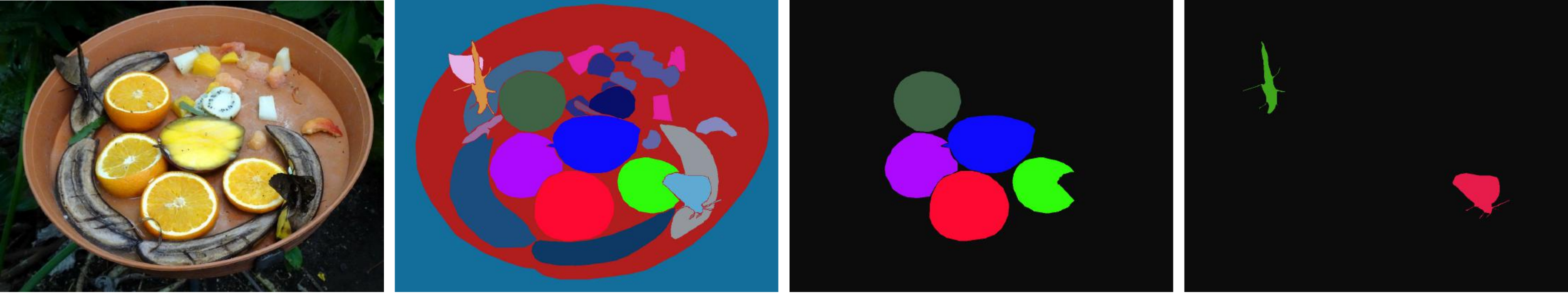}
    \put(5.8,-3.5) {\footnotesize (a) Image}
    \put(25,-3.5){\footnotesize (b) Generic object}
    \put(51,-3.5){\footnotesize (c) Salient object}
    \put(73.5,-3.5){\footnotesize (d) Concealed object}
  \end{overpic} \\
  \caption{\textbf{Task relationship.}
    Given an input image (a), we present the ground-truth for 
    (b) panoptic segmentation~\cite{kirillov2019panoptic} 
    (which detects \textbf{generic}
    objects~\cite{liu2019deep,medioni2009generic} including stuff and things),
    (c) instance level \textbf{salient} object detection
    \cite{li2017instance,Fan2021SOC}, and (d) the proposed 
    \textbf{concealed} object detection task, 
    where the goal is to detect objects that have a similar pattern 
    to the natural \Rev{environment.}
    In this example, the boundaries of the two butterflies are blended 
    with the bananas, making them difficult to identify.
  }\label{fig:differentTask}
\end{figure}

In \figref{fig:differentTask}, we present examples of generic, salient, 
and concealed object detection.
The high intrinsic similarities between the targets and non-targets make 
COD far more challenging than traditional object segmentation/detection~\cite{zhao2019object,Fan2021SOC,zhao2019EGNet}.
Although it has gained increased attention recently, 
studies on COD still remain scarce, 
mainly due to the lack of a sufficiently large dataset and a standard benchmark
like Pascal-VOC~\cite{everingham2010pascal}, 
ImageNet~\cite{deng2009imagenet}, MS-COCO~\cite{lin2014microsoft}, 
ADE20K~\cite{zhou2017scene}, and DAVIS~\cite{perazzi2016benchmark}.

\begin{figure*}[t!]
  \centering
  \includegraphics[width=\linewidth]{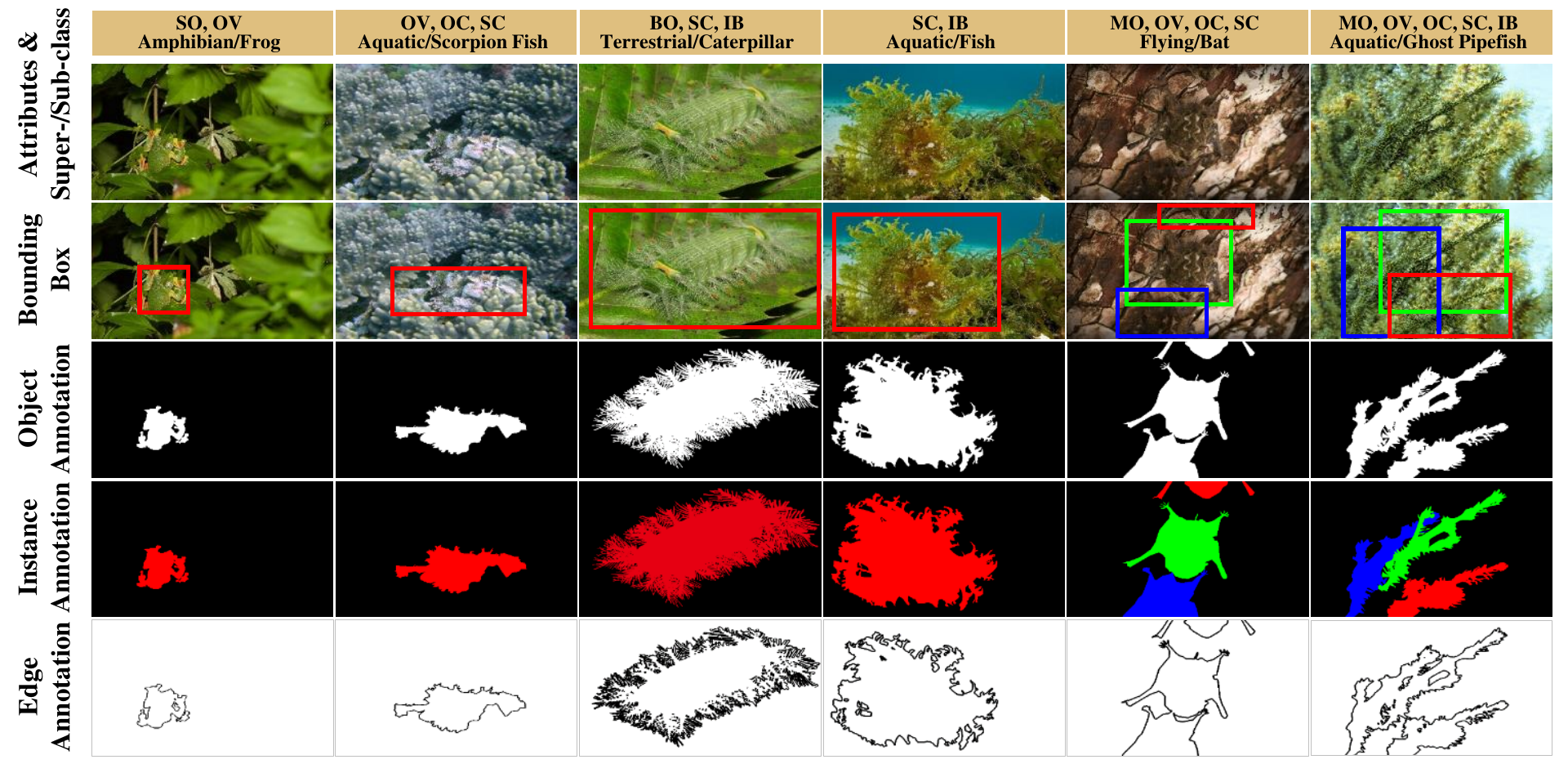}\\
  \vspace{-8pt}
  \caption{\textbf{Annotation diversity in the proposed~\ourdataset~dataset.} 
    Instead of only providing coarse-grained object-level annotations 
    like in previous works, we offer six different annotations for each image, 
    which include attributes and categories ($1^{st}$ row),  
    bounding boxes ($2^{nd}$ row), object annotation ($3^{rd}$ row), 
    instance annotation ($4^{th}$ row), and edge annotation ($5^{th}$ row).
  }\label{fig:AnnnotationDiversity}
\end{figure*}

In this paper, we present the first complete study for the 
concealed object detection task using deep learning, 
bringing a novel view to object detection from a concealed perspective. 

\subsection{Contributions}
Our main contributions are as follows:

\begin{itemize}

\item [1)] \textbf{COD10K Dataset.} 
With the goal mentioned above, we carefully assemble \ourdataset, 
a large-scale concealed object detection dataset. 
Our dataset contains 10,000 images covering 78 object categories, such as 
\emph{terrestrial}, \emph{amphibians}, \emph{flying}, \emph{aquatic}, \etc. 
All the concealed images are \emph{hierarchically annotated} with category, 
bounding-box, object-level, 
and instance-level labels (\figref{fig:AnnnotationDiversity}), 
benefiting many related tasks, such as object proposal, localization, 
semantic edge detection, 
transfer learning~\cite{zamir2018taskonomy}, 
domain adaption~\cite{saenko2010adapting}, \etc.
Each concealed image is assigned \emph{challenging attributes} 
(\eg, shape complexity-SC, indefinable boundaries-IB, occlusions-OC) 
found in the real-world and \emph{matting-level}~\cite{zhang2019late} 
labeling (which takes $\sim$60 minutes per image).
These high-quality labels could help provide deeper insight into 
the performance of models.

\item [2)] \textbf{COD Framework.} 
We propose a simple but efficient framework, named \textbf{\ournewmodel}
(\textbf{S}earch \textbf{I}dentification \textbf{Net}).
Remarkably, the overall training time of \ournewmodel~takes 4 hours and 
it achieves the new \sArt (SOTA) on all existing COD datasets, suggesting 
that it could offer a potential solution to concealed object detection. 
Our network also yield several interesting findings 
(\eg, search and identification strategy is suitable for COD), 
making various potential applications more feasible. 

\item [3)] \textbf{COD Benchmark.} 
Based on the collected \ourdataset~and previous datasets
\cite{2018Animal,le2019anabranch}, 
we offer a rigorous evaluation of \baselineN~SOTA baselines,
making ours the largest COD study. 
We report baselines in two scenarios, \ie, \textit{super-class} and 
\textit{sub-class}. 
We also track the community's progress via an online benchmark (\Rev{\url{http://dpfan.net/camouflage/}}). 

\item [4)] \textbf{Downstream Applications.} 
To further support research in the field, 
we develop an online demo (\url{http://mc.nankai.edu.cn/cod}) to enable other researchers to test 
their scenes easily. 
In addition, we also demonstrate several potential applications 
such as medicine, manufacturing, agriculture, art, \etc.

\item [5)] \textbf{Future Directions.} 
Based on the proposed \ourdataset, 
we also discuss ten promising directions for future research. 
We find that concealed object detection is still far from \Rev{being} solved, 
leaving large room for improvement. 

\end{itemize}

This paper is based on and extends our conference version~\cite{fan2020camouflaged} in terms of several aspects. 
First, we provide a more detailed analysis of our \ourdataset, 
including the taxonomy, statistics, annotations, and resolutions. 
\Rev{Secondly}, we improve the performance our \ournewmodel~model by introducing 
neighbor connection decoder (NCD) and group-reversal attention (GRA). 
\Rev{Thirdly}, we conduct extensive experiments to validate the effectiveness of 
our model, and provide several ablation studies for the different modules 
within our framework.
Fourth, we provide an exhaustive super-class and sub-class benchmarking 
and a more insightful discussion regarding the novel COD task.
Last but not least, based on our benchmark results, 
we draw several important conclusions and 
highlight several promising future directions,  
such as concealed object ranking, concealed object proposal, 
concealed instance segmentation.

\section{Related Work}\label{sec:RelatedWOrks}

In this section, we briefly review closely related works. 
Following~\cite{zhao2019object}, we roughly divide object detection into 
three categories: generic, salient, and concealed object detection. 

\begin{table*}[t!]
  \centering
  \renewcommand{\arraystretch}{0.9}
  \setlength\tabcolsep{5.1pt}
  \caption{\textbf{Summary of COD datasets, showing that 
    \ourdataset~offers much richer annotations and benefits many tasks.}
    Att.: Attribute label.
    BBox.: Bounding box label.
    Ml.: Alpha matting-level annotation~\cite{zhang2019late}.
    Ins.: Instance-level label.
    Cate.: Category label.
    Obj.: Object-level label.
    Loc.: Location.
    Det.: Detection.
    Cls.: Classification.
    WS.: Weak Supervision.
    InSeg. Instance Segmentation.
  }\label{tab:DatasetSummary}
  \vspace{-10pt}
  \begin{tabular}{r|crr|cccccc|rr|ccccc} \hline
   & \multicolumn{3}{c|}{Statistics} 
   & \multicolumn{6}{c|}{Annotations} 
   & \multicolumn{2}{c|}{Data Split}  
   & \multicolumn{5}{c}{ Tasks }\\
   \cline{2-17}
   Dataset& Year & \#Img. & \#Cls. & Att. & BBox. & Ml. & Ins. & Cate. & Obj. & \#Training & \#Testing & Loc. & Det. &  Cls. & WS. & InSeg.\\
  \hline
  \textit{CHAMELEON}~\cite{2018Animal}      & 2018 & 76     & N/A & $\times$ & $\times$ & $\times$ & $\times$ & $\times$ & \checkmark &  0 & 76 &\checkmark & \checkmark & $\times$ & $\times$ & $\times$\\
  
  
  \textit{CAMO-COCO}~\cite{le2019anabranch} & 2019 & 2,500  & 8   &  \checkmark & $\times$  & $\times$ & $\times$ & $\times$ &\checkmark& 1,250 & 1,250 &\checkmark &  \checkmark  & $\times$ & $\times$ & $\times$\\
  \rowcolor{mygray}
  \textbf{\ourdataset~(OUR)}               & 2020 & \textbf{10,000} &\textbf{78}& \checkmark &\checkmark & \checkmark &\checkmark &\checkmark &\checkmark & \textbf{6,000} & \textbf{4,000} &\checkmark & \checkmark &\checkmark &\checkmark & \checkmark\\
  \hline
  \end{tabular}
\end{table*}


\myPara{Generic Object Segmentation (GOS).}
One of the most popular directions in computer vision is generic object 
segmentation~\cite{shotton2006textonboost,liu2010sift,
everingham2015pascal,kirillov2019panoptic}.
Note that generic objects can be either salient or concealed.
Concealed objects can be seen as difficult cases of generic objects. 
Typical GOS tasks include semantic segmentation and panoptic segmentation 
(see \figref{fig:differentTask} b). 

\myPara{Salient Object Detection (SOD).}
This task aims to identify the most attention-grabbing objects
in an image and then segment their pixel-level silhouettes
\cite{itti1998model,ChengPAMI,GaoEccv20Sal100K}.
The flagship products that make use of SOD technology~\cite{HouPami19Dss} are 
Huawei's smartphones, 
which employ SOD \cite{HouPami19Dss} to create what they call ``AI Selfies''. 
Recently, Qin \etal~applied the SOD algorithm~\cite{qin2021BAS} to two (near)
commercial applications: AR COPY \& PASTE\footnote{
\url{https://github.com/cyrildiagne/ar-cutpaste}} 
and OBJECT CUT\footnote{\url{https://github.com/AlbertSuarez/object-cut}}.
These applications have already drawn great attention 
(12K github stars) and have important real-world impacts.

Although the term ``salient'' is essentially the opposite of ``concealed'' 
(\emph{standout} vs. \emph{immersion}),
salient objects can nevertheless provide important information for COD,
\eg, images containing salient objects can be used as the negative samples.
Giving a complete review on SOD is beyond the scope of this work. 
We refer readers to recent survey and benchmark papers
\cite{Fan2021SOC,borji2015salient,wang2019salient,BorjiCVM2019} 
for more details. Our online benchmark is publicly available at: 
\url{http://dpfan.net/socbenchmark/}.

\myPara{Concealed Object Detection (COD).}
Research into COD, which has had a tremendous impact on advancing our knowledge of 
visual perception, has a long and rich history in biology and art.
Two remarkable studies on concealed animals from Abbott Thayer
\cite{thayer1909concealing} and Hugh Cott~\cite{cott1940adaptive}
are still hugely influential. 
The reader can refer to the survey by Stevens~\etal~\cite{stevens2008animal} 
for more details on this history. 
There are also some concurrent works~\cite{zhai2021Mutual,Lyu2021Mutual,mei2021Ming} 
that are accepted after this submission.

\textit{COD Datasets.}
CHAMELEON~\cite{2018Animal} is an unpublished dataset that has only 76 images 
with manually annotated object-level ground-truths (GTs).
The images were collected from the Internet via the Google search engine using
``concealed animal'' as a keyword.
%
%
Another contemporary dataset is CAMO~\cite{le2019anabranch}, which has
2.5K images (2K for training, 0.5K for testing) covering eight categories.
It has two sub-datasets, CAMO and MS-COCO, each of which contains 1.25K images.
Unlike existing datasets, the goal of our \ourdataset~is to provide 
a more challenging, higher quality, and more densely annotated dataset.
\ourdataset~is the largest concealed object detection dataset
so far, containing 10K images (6K for training, 4K for testing).
See \tabref{tab:DatasetSummary} for details.

\textit{Types of Camouflage.}
Concealed images can be roughly split into two types:
those containing natural camouflage and those with artificial camouflage.
Natural camouflage is used by animals (\eg, insects, sea horses, 
and cephalopods) as a survival skill to avoid recognition by predators. 
In contrast, artificial camouflage is usually used in art design/gaming to 
hide information, occurs in products during the manufacturing process 
(so-called surface defects~\cite{tabernik2020segmentation}, 
defect detection~\cite{he2020an,dong2020pga}), 
or appears in our daily life (\eg, transparent objects
\cite{kalra2020deep,xu2015transcut,xie2020segmenting}).

\textit{COD Formulation.}
Unlike class-aware tasks such as semantic segmentation, 
concealed object detection is a class-agnostic task. 
Thus, the formulation of COD is simple and easy to define.
Given an image, the task requires a \emph{concealed object detection algorithm} 
to assign each pixel $i$ a label $Label_i \in$ \{0,1\}, 
where $Label_i$ denotes the binary value of pixel $i$.
A label of 0 is given to pixels that do not belong to the concealed objects,
while a label of 1 indicates that a pixel is fully assigned to the concealed objects.
We focus on object-level concealed object detection, 
leaving concealed instance detection to our future work.

\begin{figure}[b]
  \centering
  \begin{overpic}[width=\columnwidth]{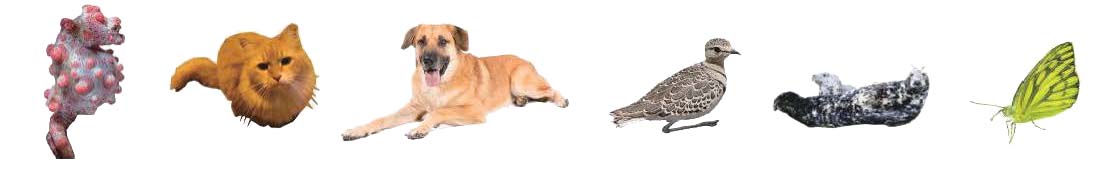}
    \put(1,-1){sea-horse}
    \put(22,-1){cat}
    \put(40,-1){dog}
    \put(60,-1){bird}
    \put(72,-1){sea-lion}
    \put(87,-1){butterfly}
  \end{overpic}\\
  \vspace{-6pt}
  \caption{\textbf{Examples of sub-classes.}
	Please refer to \supp{supplementary materials} for other sub-classes.
  }\label{fig:SubClassExample}
\end{figure}

\begin{figure*}[t!]
  \centering
  \includegraphics[width=\textwidth]{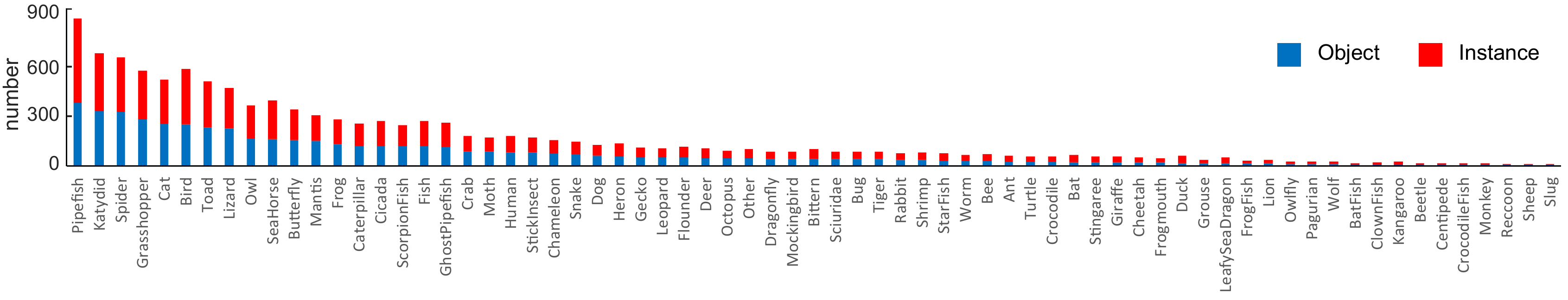}
  \vspace{-25pt}
  \caption{\textbf{Object and instance distributions of each concealed category in the \ourdataset.} 
    \ourdataset~consists of 5,066 concealed images from 69 categories. 
    Zoom in for best view.
  }\label{fig:ObjInstNumberHist}
\end{figure*}

\section{COD10K Dataset}\label{sec:CODdataset}

The emergence of new tasks and datasets
\cite{cordts2016cityscapes,zhou2017scene,neuhold2017mapillary} 
has led to rapid progress in various areas of computer vision.
For instance, ImageNet~\cite{russakovsky2015imagenet} revolutionized 
the use of deep models for visual recognition.
With this in mind, our goals for studying and developing a dataset for COD are:
(1) to provide a new challenging object detection task from the concealed 
perspective,
(2) to promote research in several new topics, and
(3) to spark novel ideas.
Examples from \ourdataset~are shown in \figref{fig:COD10KExample}. 
We will provide the details on \ourdataset~in terms of three key aspects
including image collection, professional annotation, and dataset features 
and statistics.

\begin{figure}[t!]
  \centering
  \includegraphics[width=\columnwidth]{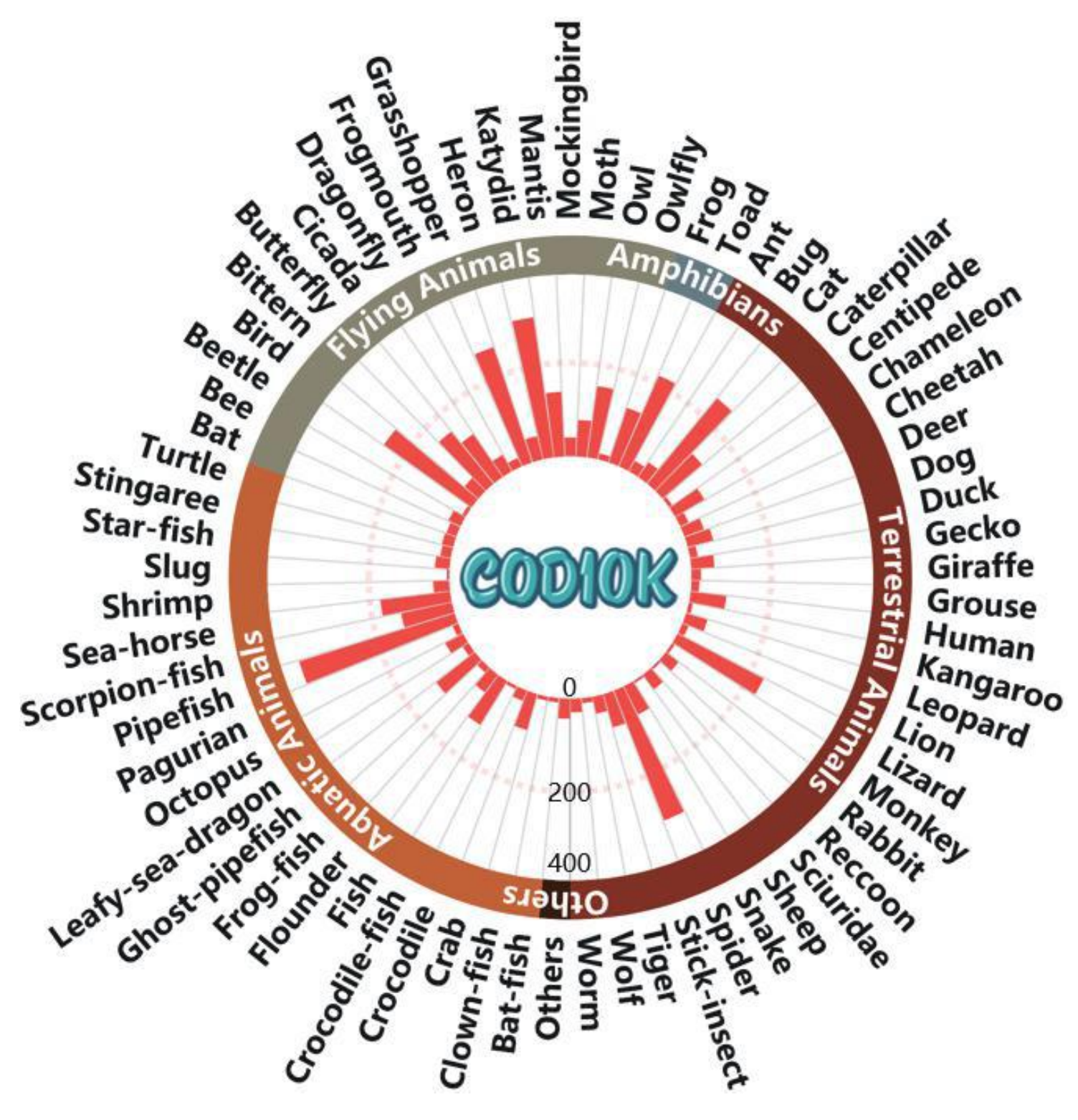}\\
  \vspace{-18pt}
  \caption{\textbf{Taxonomic system.} We illustrate the histogram distribution 
    for the 69 concealed categories in our \ourdataset.
  }\label{fig:SubClassSystem}
\end{figure}

\subsection{Image Collection}

As discussed in~\cite{perazzi2016benchmark,wang2018revisiting,Fan2021SOC},
the quality of annotation and size of a dataset are determining factors for its
lifespan as a benchmark.
To this end, \ourdataset~contains 10,000 images (5,066 concealed, 
3,000 background, 1,934 non-concealed), 
divided into 10 super-classes (\ie, flying, aquatic, terrestrial, 
amphibians, other, sky, vegetation, indoor, ocean, and sand), 
and 78 sub-classes (69 concealed, 9 non-concealed) 
which were collected from multiple photography websites.

Most concealed images are from \Rev{Flickr} and have been applied for academic
use with the following keywords:
\emph{concealed animal}, \emph{unnoticeable animal}, \emph{concealed fish},
\emph{concealed butterfly}, \emph{hidden wolf spider}, \emph{walking stick},
\emph{dead-leaf mantis}, \emph{bird}, \emph{sea horse}, \emph{cat},
\emph{pygmy seahorses}, \etc.~(see \figref{fig:SubClassExample})
The remaining concealed images (around 200 images) come from other websites,
including Visual Hunt, Pixabay, Unsplash, Free-images, \etc.,
which release public-domain stock photos, free from copyright and loyalties.
To avoid selection bias~\cite{Fan2021SOC}, 
we also collected 3,000 salient images from Flickr. 
To further enrich the negative samples, 1,934 non-concealed
images, including \emph{forest}, \emph{snow}, \emph{grassland}, \emph{sky}, \emph{seawater}
and other categories of background scenes, were selected from the Internet.
For more details on the image selection scheme, we refer to Zhou \etal~\cite{zhou2017places}.

\subsection{Professional Annotation}

Recently released datasets
\cite{wang2018revisiting,Fan2019D3Net,damen2018scaling} 
have shown that establishing a taxonomic system is crucial when creating 
a large-scale dataset.
Motivated by~\cite{mo2019partnet}, our annotations (obtained via crowdsourcing) 
are hierarchical (category $\rightarrowtail$ bounding box $\rightarrowtail$
attribute $\rightarrowtail$  object/instance).


$\bullet$ \emph{Categories.}
As illustrated in \figref{fig:SubClassSystem}, 
we first create five super-class categories.
Then, we summarize the 69 most frequently appearing sub-class categories 
according to our collected data.
Finally, we label the sub-class and super-class of each image.
If the candidate image does not belong to any established category,
we classify it as `other'.

\begin{figure}[t!]
  \centering
  \includegraphics[width=\columnwidth]{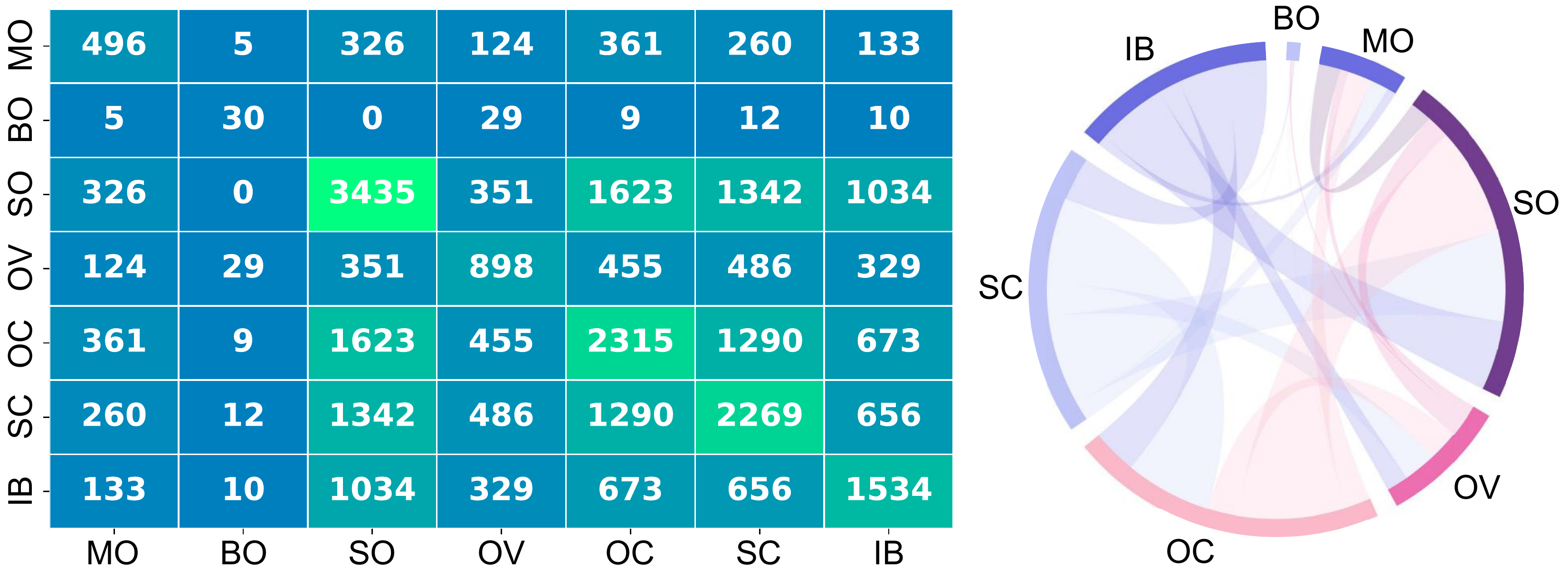}\\
  \vspace{4pt}
  \footnotesize
  \renewcommand{\arraystretch}{0.9}
 \begin{tabular*}{\linewidth}{cl} \hline \toprule
  Attr. & Description\\
  \midrule
  \textbf{MO} & \emph{Multiple Objects.} Image contains at least two objects.\\
  \textbf{BO} & \emph{Big Object.} Ratio ($\tau_{bo}$) between object area 
                and image area $\geq$0.5.\\
  \textbf{SO} & \emph{Small Object.} Ratio ($\tau_{so}$) between object area 
                and image area $\leq$0.1.\\
  \textbf{OV} & \emph{Out-of-View.} Object is clipped by image boundaries.\\
  \textbf{OC} & \emph{Occlusions.} Object is partially occluded.\\
  \textbf{SC} & \emph{Shape Complexity.} Object contains thin parts 
                (\eg, animal foot).\\
  \textbf{IB} & \emph{Indefinable Boundaries.} 
                The foreground and background areas\\
              & around the object have similar colors ($\chi^2$ distance 
                $\tau_{gc}$ between\\
              & RGB histograms less than 0.9).\\
  \hline \toprule
 \end{tabular*}
  \vspace{-8pt}
  \caption{\textbf{Attribute distribution.}
    Top-left: Co-attributes distribution over \ourdataset. 
    The number in each grid indicates the total number of images.
    Top-right: Multi-dependencies among these attributes.
    A larger arc length indicates a higher probability of one attribute 
    correlating to another.
    Bottom: attribute descriptions. 
    Examples could be found in the first row of 
    \figref{fig:AnnnotationDiversity}.
  }\label{fig:attributeDistribution}
\end{figure}

\begin{figure*}[t!]
  \centering
  \includegraphics[width=.95\textwidth]{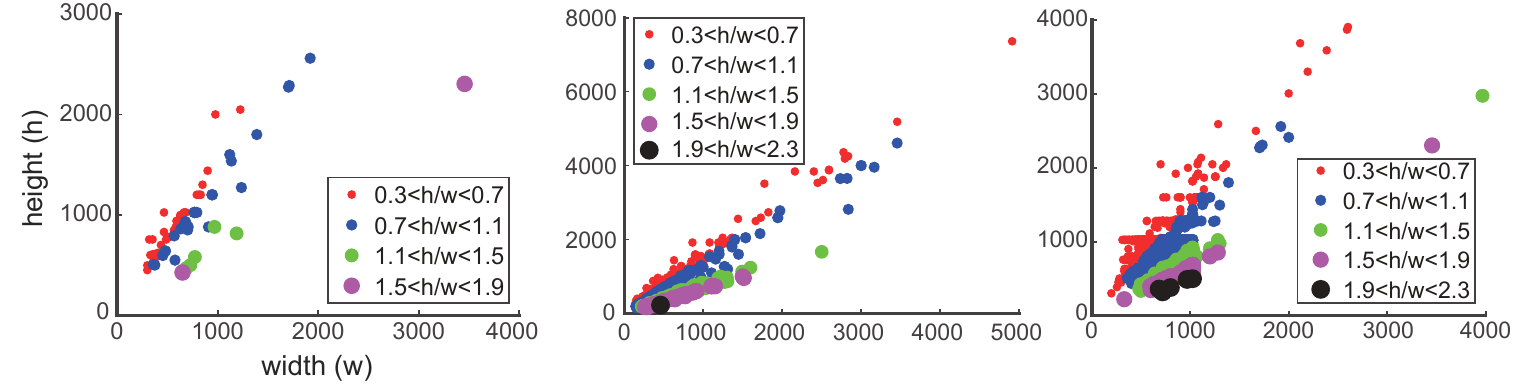}\\
  \vspace{-12pt}
  \caption{\textbf{Image resolution \Rev{(unit for the axis: pixel)} distribution of COD datasets.}
    From left to right: 
    CHAMELEON~\cite{2018Animal},
    CAMO-COCO~\cite{le2019anabranch} and \ourdataset~datasets.
  }\label{fig:Resolution3Datasets}
\end{figure*}

$\bullet$ \emph{Bounding boxes.}
To extend \ourdataset~for the concealed object proposal task,
we also carefully annotate the bounding boxes for each image.

$\bullet$ \emph{Attributes.}
In line with the literature~\cite{Fan2021SOC,perazzi2016benchmark}, 
we label each concealed image 
with highly challenging attributes faced in natural scenes,
\eg, \emph{occlusions}, \emph{indefinable boundaries}.
Attribute descriptions 
and the co-attribute distribution is shown in 
\figref{fig:attributeDistribution}.

\begin{figure}[t!]
  \centering
  \includegraphics[width=.98\columnwidth]{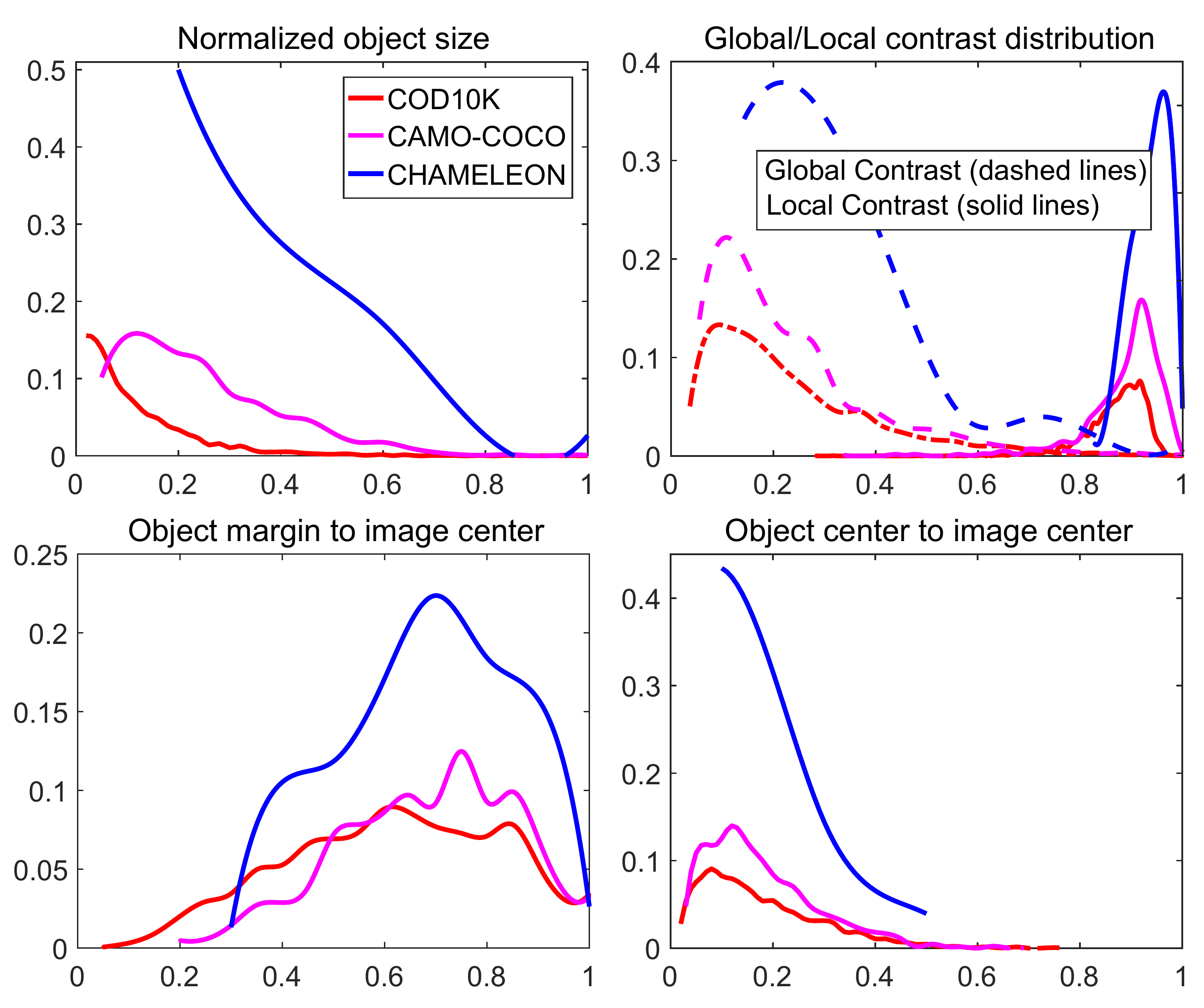}\\
  \vspace{-10pt}
  \caption{\textbf{Comparison between the proposed \ourdataset~and 
    existing datasets.}
    \ourdataset~has smaller objects (top-left),
    contains more difficult conceale (top-right),
    and suffers from less center bias (bottom-left/right).
  }\label{fig:DistributionStatistics}
\end{figure}

\begin{figure}[t!]
  \centering
  \includegraphics[width=\columnwidth]{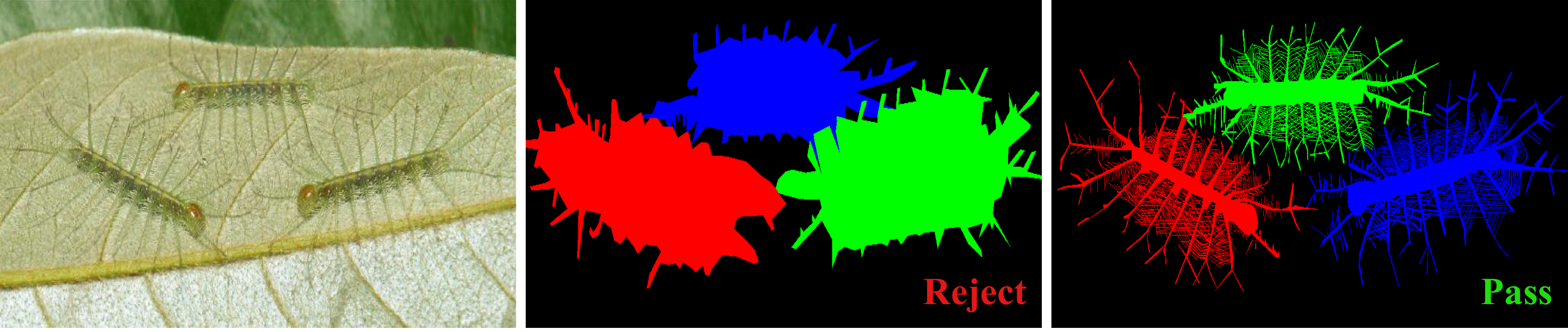}\\
  \vspace{-10pt}
  \caption{\textbf{High-quality annotation.}
	The annotation quality is close to the existing matting-level~\cite{zhang2019late} annotation.
  }\label{fig:highqualityAnnotation}
\end{figure}

$\bullet$ \emph{Objects/Instances.}
We stress that existing COD datasets focus exclusively on
object-level annotations (\tabref{tab:DatasetSummary}).
However, being able to parse an object into its instances is
important for computer vision researchers to be able to edit
and understand a scene.
To this end, we further annotate objects at an instance-level,
like COCO~\cite{lin2014microsoft}, 
resulting in 5,069 objects and 5,930 instances.

\subsection{Dataset Features and Statistics}\label{sec:datasetFeatures}
We now discuss the proposed dataset and provide some statistics. 

$\bullet$ \emph{Resolution distribution.}
As noted in~\cite{zeng2019towards}, high-resolution data
provides more object boundary details for model training and
yields better performance when testing.
\figref{fig:Resolution3Datasets} presents the resolution distribution of 
\ourdataset, which includes a large number of Full HD 1080p resolution images.

$\bullet$ \emph{Object size.}
Following~\cite{Fan2021SOC}, we plot the normalized (\Rev{\ie, related to image areas}) object size in
\figref{fig:DistributionStatistics} (top-left), \ie, the size distribution from
0.01\%$\sim$ 80.74\% (avg.: 8.94\%), showing a broader range compared to
CAMO-COCO, and CHAMELEON.

$\bullet$ \emph{Global/Local contrast.}
To evaluate whether an object is easy to detect,
we describe it using the global/local contrast strategy~\cite{li2014secrets}.
\figref{fig:DistributionStatistics} (top-right) shows that objects
in \ourdataset~are more challenging than those in other datasets.

$\bullet$ \emph{Center bias.}
This commonly occurs when taking a photo, as
humans are naturally inclined to focus on the center of a scene.
We adopt the strategy described in~\cite{Fan2021SOC} to analyze this bias.
\figref{fig:DistributionStatistics} (bottom-left/right) shows that our 
\ourdataset~dataset suffers from less center bias than others.

$\bullet$ \emph{Quality control.}
To ensure high-quality annotation, we invited three viewers to participate in
the labeling process for 10-fold cross-validation.
\figref{fig:highqualityAnnotation} shows examples that were passed/rejected.
This matting-level annotation costs $\sim$ 60 minutes per image on average.

\begin{figure}[t!]
  \centering
  \includegraphics[width=\columnwidth]{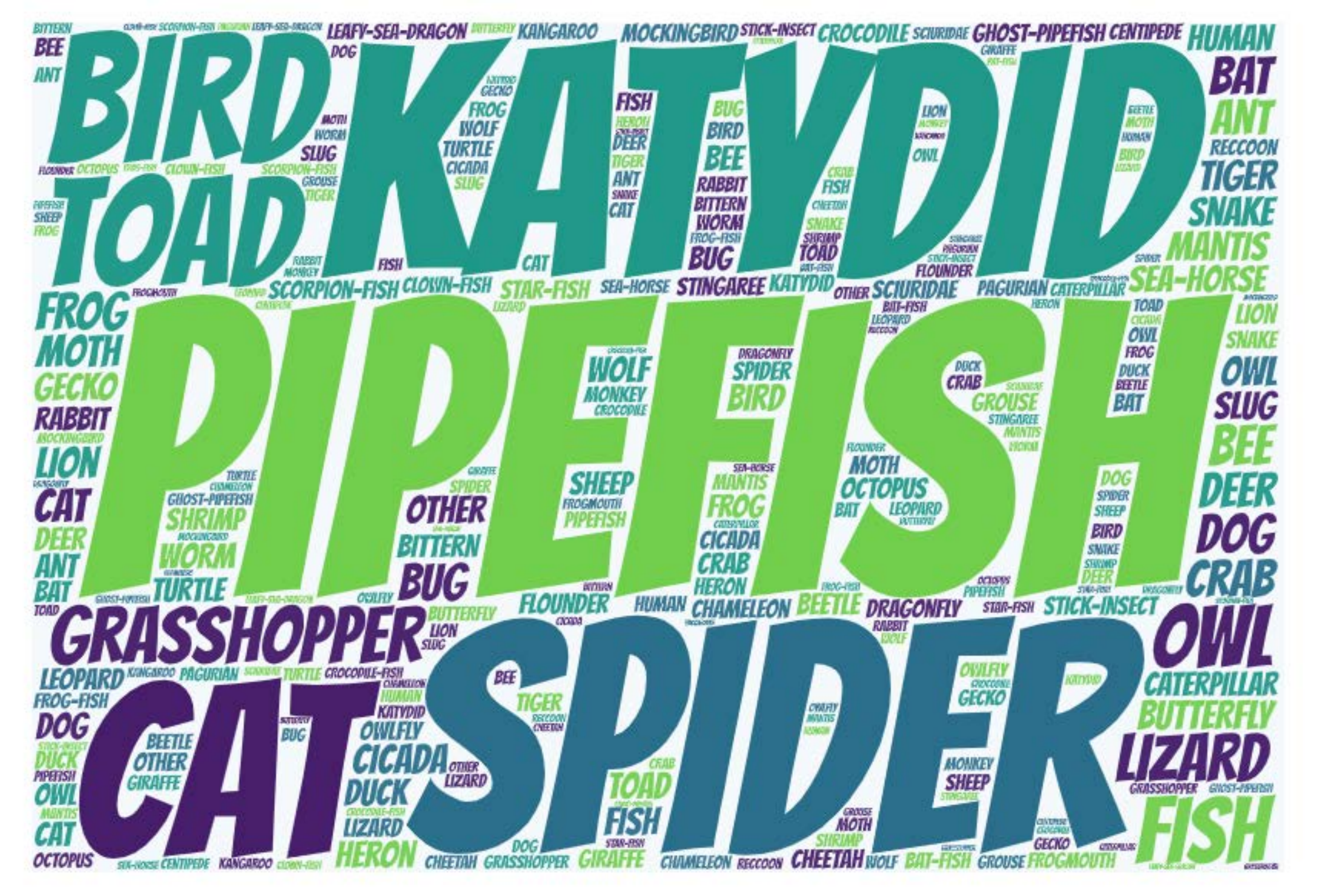}\\
  \vspace{-15pt}
  \caption{\textbf{Word cloud distribution.}
	The size of a specific word is proportional to the ratio 
	of that keyword.
  }\label{fig:Wordcloud}
\end{figure}

$\bullet$ \emph{Super/Sub-class distribution.}
\ourdataset~includes five concealed super-classes (\ie, \emph{terrestrial}, 
\emph{atmobios}, \emph{aquatic}, \emph{amphibian}, \emph{other}) and
69 sub-classes (\eg, \emph{bat-fish}, \emph{lion}, \emph{bat}, \emph{frog}, 
\etc). 
Examples of the word cloud and object/instance number for various categories are
shown in \figref{fig:ObjInstNumberHist} \& \figref{fig:Wordcloud}, respectively.

\begin{figure}[t!]
  \centering
  \includegraphics[width=.98\columnwidth]{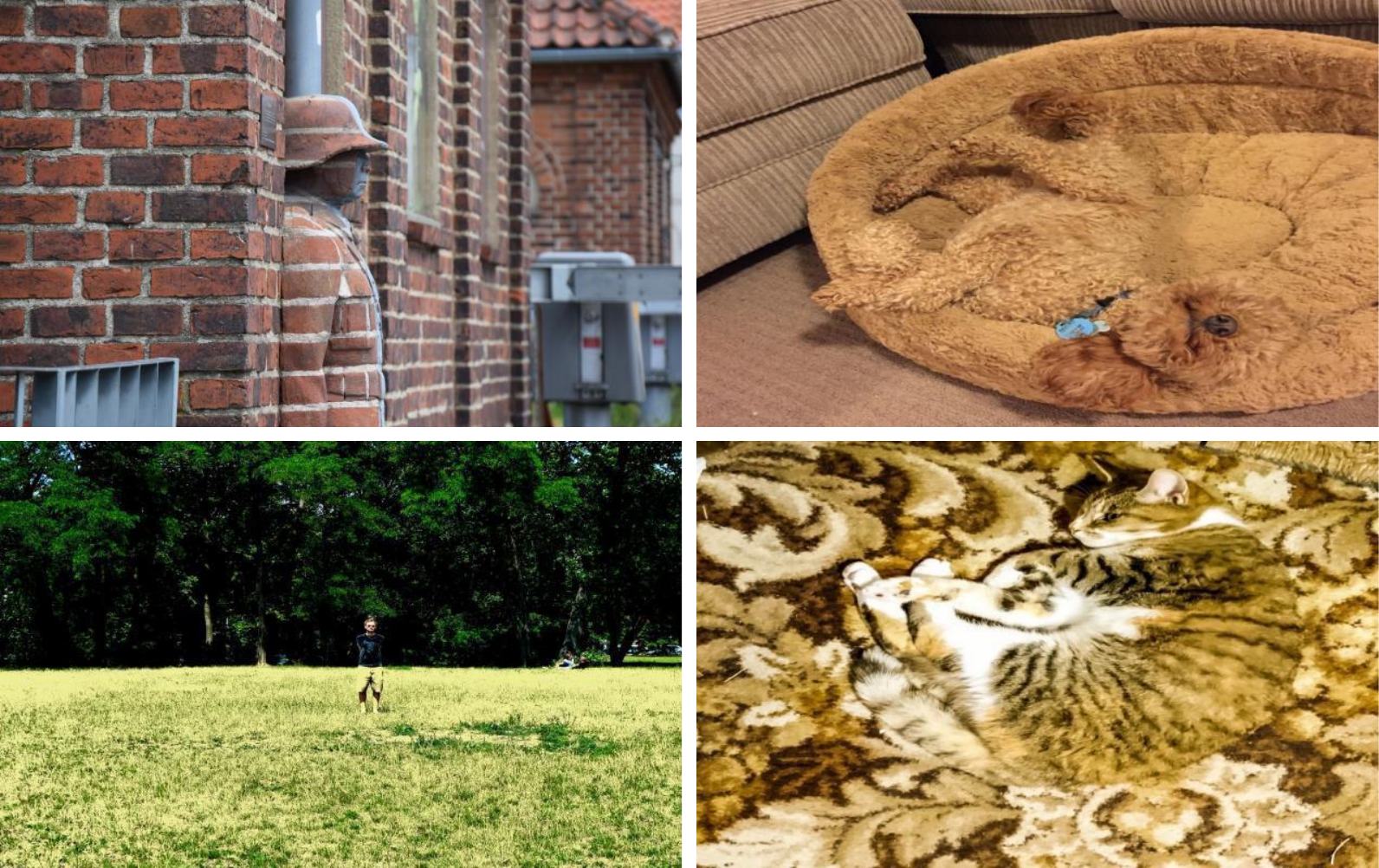}\\
  \vspace{-10pt}
  \caption{\textbf{Diverse types of concealed objects in our~\ourdataset.} 
	For instance, concealed human in art ($1^{st}$ column), and 
    concealed animals ($2^{nd}$ column) in our daily life.
  }\label{fig:DiverseType}
\end{figure}

$\bullet$ \emph{Dataset splits.}
To provide a large amount of training data for deep learning algorithms, 
our \ourdataset~is split into 6,000 images for training and 4,000 for testing, 
randomly selected from each sub-class.

$\bullet$ \emph{Diverse concealed objects.}
In addition to the general concealed patterns, such as those in
\figref{fig:COD10KExample}, our dataset also includes various 
other types of concealed objects, such as
concealed body paintings and conceale in daily life 
(see \figref{fig:DiverseType}).

\begin{figure*}[t!]
  \centering
  \includegraphics[width=\textwidth]{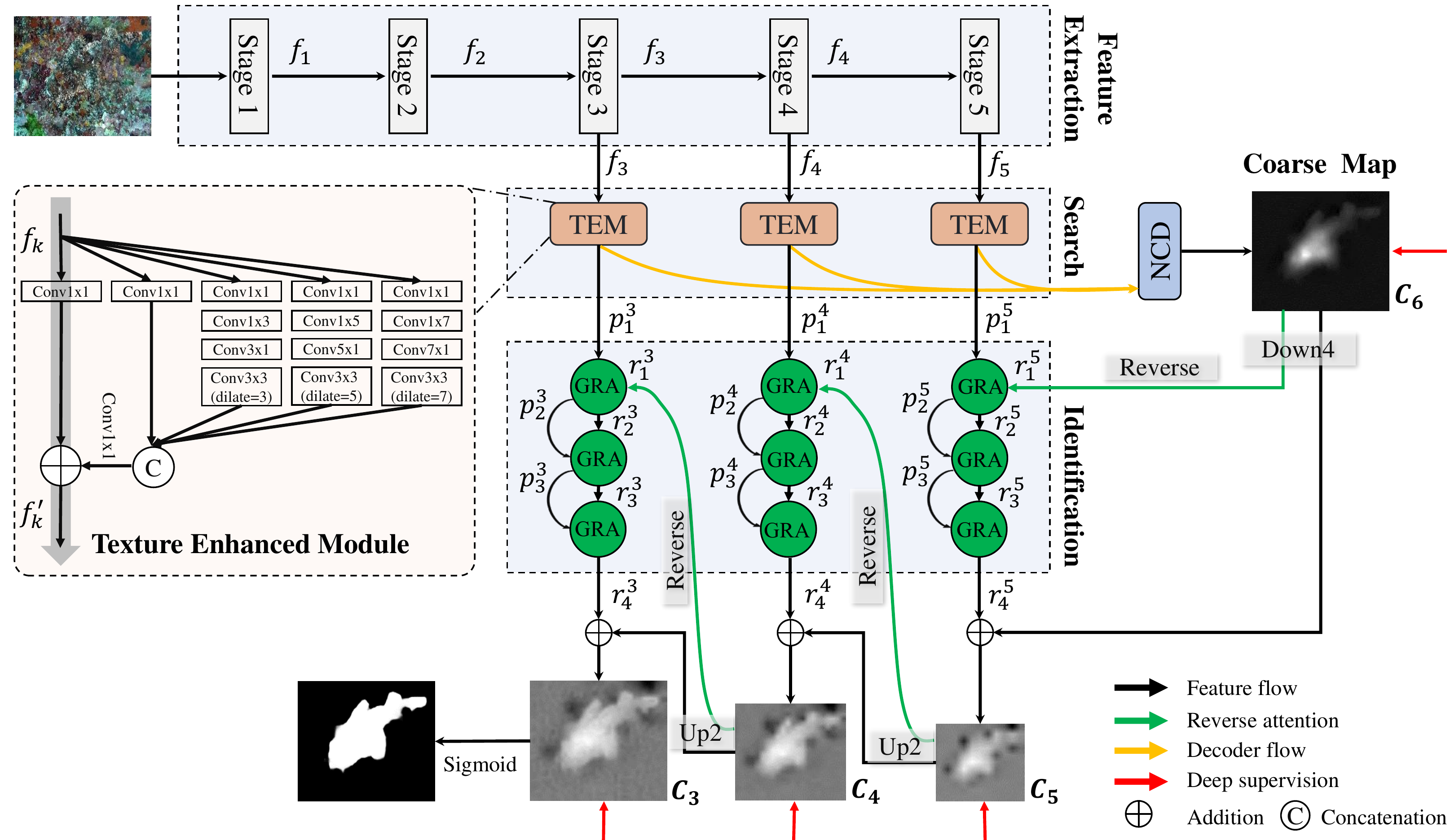}\\
  \vspace{-12pt}
  \caption{\textbf{Pipeline of our~\ournewmodel~framework}. 
    It consists of three main components: the texture enhanced module (TEM), 
    neighbor connection decoder (NCD), and group-reversal attention (GRA).
    The TEM is introduced to mimic the textural structure of receptive fields 
    in the human visual system.
    The NCD is responsible for locating the candidates with the assistance 
    of the TEM.
    The GRA blocks reproduce the identification stages of animal predation. 
    Note that $f'_k$ = $p_1^k$.
  }\label{fig:Framework}
\end{figure*}

\section{COD Framework}\label{sec:ourmodel}

\subsection{Network Overview}
\figref{fig:Framework} illustrates the overall concealed object detection 
framework of the proposed \ournewmodel~(Search Identification Network). 
Next, we explain our motivation and introduce the network overview.

\myPara{Motivation.}
Biological studies~\cite{hall2013camouflage} have shown that, when hunting, 
a predator will first judge whether a potential prey exists, \ie, 
it will \emph{search} for a prey.
Then, the target animal can be \emph{identified}; and, finally, 
it can be \emph{caught}.

\myPara{Introduction.}
Several methods~\cite{qin2019basnet,xu2017deep} have shown that satisfactory 
performance is dependent on the re-optimization strategy (\ie, coarse-to-fine), 
which is regarded as the composition of multiple sub-steps.
This also suggests that decoupling the complicated targets can break the
performance bottleneck.
%
Our \ournewmodel~model consists of the first two stages of hunting, 
\ie, search and identification.
Specifically, the former phase (\secref{sec:search_phase}) is responsible for 
searching for a concealed object, 
while the latter one (\secref{sec:identification_phase}) is then 
used to precisely detect the concealed object in a cascaded manner.

Next, we elaborate on the details of the three main modules, including 
\textit{a)} the texture enhanced module (\textbf{TEM}), 
which is used to capture fine-grained textures with the enlarged context cues; 
\textit{b)} the neighbor connection decoder (\textbf{NCD}), 
which is able to provide the location information; and 
\textit{c)} the cascaded group-reversal attention (\textbf{GRA}) blocks, 
which work collaboratively to refine the coarse prediction from 
the deeper layer.

\subsection{Search Phase}\label{sec:search_phase}

\myPara{Feature Extraction.}
For an input image $\mathbf{I}\!\in\!\mathbb{R}^{W\!\times\! H \!\times\!3}$, 
a set of features $f_k, k \in \{ 1,2,3,4,5 \}$ is extracted from 
Res2Net-50~\cite{gao2019res2net} (\Rev{removing the top three layers, \ie, `average pool', `1000-d fc', and `softmax'}).
Thus, the resolution of each feature 
$f_k$ is ${H/2^k \times W/2^k}, k \in \{ 1,2,3,4,5 \}$, 
covering diversified feature pyramids from high-resolution, 
weakly semantic to low-resolution, strongly semantic.

\myPara{Texture Enhanced Module (TEM).}
Neuroscience experiments have verified that, in the human visual system, 
a set of various sized population receptive fields helps to 
highlight the area close to the retinal fovea, 
which is sensitive to small spatial shifts~\cite{liu2018receptive}.
This motivates us to use the TEM~\cite{wu2019cascaded} to incorporate 
more discriminative feature representations during the searching stage 
(usually in a small/local space).
As shown in~\figref{fig:Framework}, each TEM component includes four parallel 
residual branches $\{b_i, i = 1,2,3,4\}$ with different dilation rates 
$d \in \{1,3,5,7\}$ and a shortcut branch (gray arrow), respectively.
In each branch $b_i$, the first convolutional layer utilizes a $1\times 1$ 
convolution operation (Conv1$\times$1) to reduce the channel size to 32.
This is followed by two other layers: 
a $(2i-1)\times(2i-1)$ convolutional layer and a $3\times 3$ 
convolutional layer with a specific dilation rate $(2i-1)$ when $i>1$.
Then, the first four branches \Rev{$\{ b_i, i=1, 2, 3, 4 \}$} are concatenated and 
the channel size is reduced to $C$ via a 3$\times$3 convolution operation.
Note that we set $C=32$ in the default implementation of our network 
for time-cost trade-off.
Finally, the identity shortcut branch is added in, then the whole module 
is fed to a ReLU function to obtain the output feature $f_k'$.
%
Besides, several works (\eg, Inception-V3~\cite{szegedy2016rethinking}) 
have suggested that the standard convolution operation of size 
$(2i-1) \times (2i-1)$ can be factorized as a sequence of two steps with 
$(2i-1) \times 1$ and $1 \times (2i-1)$ kernels, 
speeding-up the inference efficiency without decreasing the 
representation capabilities.
All of these ideas are predicated on the fact that a 2D kernel with a rank 
of one is equal to a series of 
1D convolutions~\cite{jaderberg2014speeding,denton2014exploiting}.
In brief, compared to the standard receptive fields block structure~\cite{liu2018receptive}, 
TEM add one more branch with a larger dilation rate to enlarge the 
receptive field and further replace the standard convolution 
with two asymmetric convolutional layers.
For more details please refer to~\figref{fig:Framework}.

\myPara{Neighbor Connection Decoder (NCD).}
As observed by Wu~\etal~\cite{wu2019cascaded}, 
low-level features consume more computational resources due to 
their larger spatial resolutions, but contribute less to performance.
Motivated by this observation, 
we decide to aggregate only the top-three highest-level features 
(\ie, $\{f_k \in \mathbb{R}^{W/2^{k} \times H/2^{k} \times C}, k = 3,4,5 \}$) 
to obtain a more efficient learning capability, 
rather than taking all the feature pyramids into consideration.
To be specific, after obtaining the candidate features from the three previous 
TEMs, in the search phase, we need to locate the concealed object.

However, there are still two key issues when aggregating multiple 
feature pyramids; namely, how to maintain semantic consistency within a layer 
and how to bridge the context across layers.
%
Here, we propose to address these with the \textit{neighbor connection decoder} (NCD).
More specifically, we modify the partial decoder component (PDC)
\cite{wu2019cascaded} with a neighbor connection function and get three refined 
features $f_{k}^{nc}= F_{NC}(f_{k}'; \mathbf{W}_{NC}^{u})$, $k \in \{ 3,4,5 \}$ 
and $u \in \{ 1,2,3 \}$, which are formulated as:
\begin{equation}
\left\{
\begin{array}{l}
f_{5}^{nc}=f_{5}'\\
f_{4}^{nc}=f_{4}' \otimes g[\delta^{2}_{\uparrow}(f_{5}'); 
\mathbf{W}_{NC}^{1}]\\
f_{3}^{nc}=f_{3}' \otimes g[\delta^{2}_{\uparrow}(f_{4}^{nc}); 
\mathbf{W}_{NC}^{2}] \otimes g[\delta^{2}_{\uparrow}(f_{4}'); 
\mathbf{W}_{NC}^{3}]
\end{array} \right.
\end{equation}
where $g[\cdot;\mathbf{W}_{NC}^{u}]$ denotes a 3$\times$3 convolutional layer 
followed by a batch normalization operation. 
To ensure shape matching between candidate features, 
we utilize an upsampling (\eg, 2 times) operation 
$\delta^{2}_{\uparrow}(\cdot)$ before element-wise multiplication $\otimes$.
Then, we feed $f_{k}^{nc}, k \in \{ 3,4,5 \}$ into the neighbor connection 
decoder (NCD) and generate the coarse location map $\mathbf{C_6}$. 
%
%

\subsection{Identification Phase}\label{sec:identification_phase}

\begin{figure}[t!]
  \centering
  \includegraphics[width=\linewidth]{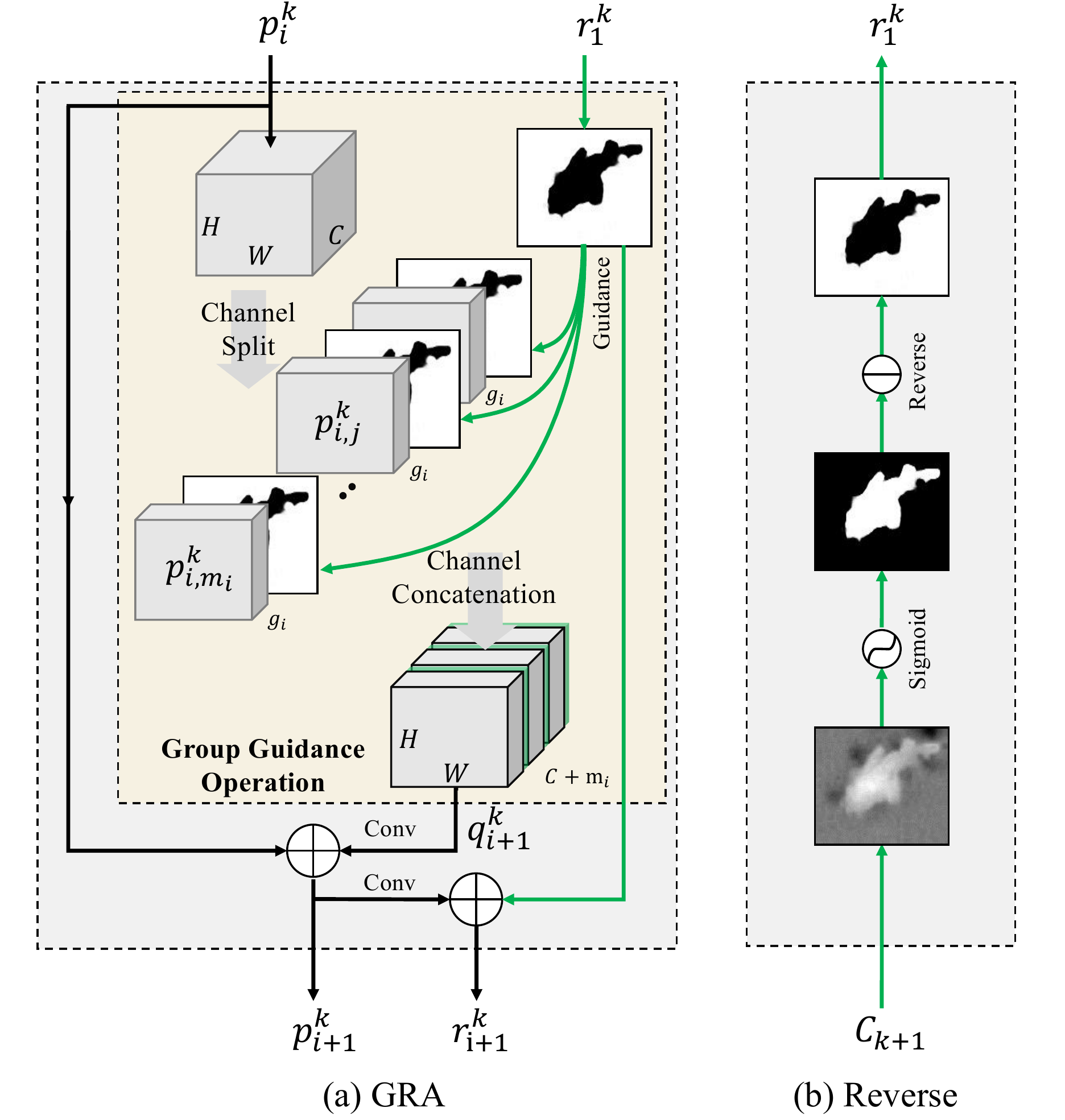}\\
  \vspace{-10pt}
  \caption{\textbf{Component details.}
    Details on the group-reversal attention (b) block $G_i^k$ 
    in the identification phase, 
    where $i$ denotes the number of GRAs in the $k$-th feature pyramids. \Rev{Note that $m_i = C/g_i$.}
  }\label{fig:GRA}
\end{figure}

\myPara{Reverse Guidance.}
As discussed in~\secref{sec:search_phase}, 
our global location map $C_6$ is derived from the three highest layers, 
which can only capture a relatively rough location of the concealed object, 
ignoring structural and textural details (see~\figref{fig:Framework}).
To address this issue, we introduce a principled strategy to mine
discriminative concealed regions 
by erasing objects~\cite{wei2017object,chen2018reverse,fan2020pranet}.
As shown in \figref{fig:GRA} (b), 
we obtain the output reverse guidance $r_1^k$ via sigmoid and reverse operation.
More precisely, we obtain the output reverse attention guidance $r_1^k$ 
by a reverse operation, which can be formulated as:
\begin{equation} \label{RA1}
r_1^k = \left\{ 
\begin{aligned} 
& \circleddash\left[ \sigma(\delta^{4}_{\downarrow}(C_{k+1})),~\mathbf{E} \right], k =5,\\ 
& \circleddash\left[ \sigma(\delta^{2}_{\uparrow}(C_{k+1})),~\mathbf{E} \right], k \in \{ 3,4 \}, 
\end{aligned} \right.
\end{equation}
where $\delta^{4}_{\downarrow}$ and $\delta^{2}_{\uparrow}$ denote a 
$\times$4 down-sampling and $\times$2 up-sampling operation, respectively. 
$\sigma(x) = 1/(1+e^{-x})$ is the \textit{sigmoid} function, 
which is applied to convert the mask into the interval [0, 1].
$\circleddash$ is a reverse operation subtracting the input from matrix 
$\mathbf{E}$, in which all the elements are $1$.

\myPara{Group Guidance Operation (GGO).}
As shown in~\cite{fan2020pranet}, 
reverse attention is used for mining complementary regions and details 
by erasing the existing estimated target regions from side-output features.
\Rev{Inspired by~\cite{chen2020progressively}}, we present a novel group-wise operation to utilize 
the reverse guidance prior more effectively.
As can be seen in \figref{fig:GRA} (a), 
the group guidance operation contains two main steps. 
First, we split the candidate features \Rev{$\{ p_i^k \in \mathbb{R}^{H/2^k \times W/2^k \times C}, k=3,4,5\}$ into multiple (\ie, $m_i = C/g_i$) groups along the channel-wise dimension, where $i=1,2,3$ and $g_i$ denotes the group size of processed features.}
Then, the guidance prior $r_1^k$ is periodically interpolated among the 
split features $p_{i,j}^{k} \in \mathbb{R}^{H/2^k \times W/2^k \times \Rev{g_i}}$, 
where $i \in \{ 1, 2, 3 \}, j \in \{ 1,\dots,\Rev{m_i} \}, k \in \{3,4,5\}$.
Thus, this operation (\ie, $q_{i+1}^{k} = \mathbf{F}^{GGO}[p_i^k, r_i^k; \Rev{m_i}]$) 
can be decoupled as two steps:
\begin{equation} \label{RA2}
  \begin{aligned}
    &\textit{Step I:}~\{ p_{i,1}^{k}, \dots, p_{i,j}^{k} , \dots , p_{i,\Rev{m_i}}^{k} \}~\gets~\mathbf{F}^S(p_i^k)\\
    &\textit{Step II:}~q_{i+1}^{k} ~\gets~\mathbf{F}^C( \{ p_{i,1}^{k}, r_1^{k} \} , \dots, \{ p_{i,j}^{k}, r_1^k \}, \dots, \{ p_{i,\Rev{m_i}}^{k}, r_1^k\} ),
  \end{aligned}
\end{equation}
where $\mathbf{F}^S$ and $\mathbf{F}^C$ indicate the channel-wise split 
and concatenation function for the candidates.
Note that $\mathbf{F}^{GGO}: p_i^k \in \mathbb{R}^{H/2^k \times W/2^k \times C} 
\rightarrow q_{i+1}^{k} \in \mathbb{R}^{H/2^k \times W/2^k \times (C+\Rev{m_i})}$, 
where $k \in \{3,4,5\}$.
In contrast, \cite{fan2020pranet} puts more emphasis on ensuring that the 
candidate features are directly multiplied by the priors, 
which may incur two issues:
\textit{a)} feature confusion due to the limited discriminative ability of 
the network, and 
\textit{b)} the simple multiplication introduces both true and false guidance 
priors and is thus prone to accumulating inaccuracies.
Compared to \cite{fan2020pranet}, our GGO can explicitly isolate the guidance 
prior and candidate feature before the subsequent refinement process.

\myPara{Group-Reversal Attention (GRA).}
Finally, we introduce the residual learning process, termed the GRA block, 
with the assistance of both the reverse guidance and group guidance operation.
According to previous studies~\cite{qin2019basnet,xu2017deep}, 
multi-stage refinement can improve performance.
We thus combine multiple GRA blocks 
(\eg, $G^k_i$, $i \in \{ 1, 2, 3 \}, k \in \{3,4,5\}$) to progressively 
refine the coarse prediction via different feature pyramids.
Overall, each GRA block has three residual learning processes:
\begin{enumerate}
\item [\textit{i)}] We combine candidate features $p_i^k$ and $r_1^k$ 
  via the group guidance operation and then use the residual stage 
  to produce the refined features $p_{i+1}^k$. 
  This is formulated as:
  \begin{equation}
    p_{i+1}^k = 
    p_i^k + g[\mathbf{F}^{GGO}[p_i^k, r_1^k; \Rev{m_i}];\mathbf{W}^v_{GRA}],
  \end{equation}
  where $\mathbf{W}^v$ denotes the convolutional layer with a 3$\times$3 
  kernel followed by batch normalization layer 
  for reducing the channel number from \Rev{$C+m_i$} to $C$. 
  Note that we only reverse the guidance prior in the first GRA block 
  (\ie, when $i=1$) in the default implementation. 
  Refer to \secref{sec:ablation_study} for detailed discussion.
    
\item [\textit{ii)}] Then, we get a single channel residual guidance:
  \begin{equation}
    r_{i+1}^k = r_1^k + g[p_{i+1}^k;\mathbf{W}^w_{GRA}],
  \end{equation}
  which is parameterized by learnable weights $\mathbf{W}^w_{GRA}$.
  
\item [\textit{iii)}] Finally, we only output the refined guidance, 
  which serves as the residual prediction. It is formulated as:
  \begin{equation}
    C_{k} = r_{i+1}^k + \delta(C_{k+1}),
  \end{equation}
  where $\delta(\cdot)$ is $\delta^2_{\uparrow}$ when $k=\{3,4\}$ 
  and $\delta^4_{\downarrow}$ when $k=5$.

\end{enumerate}

\subsection{Implementation Details}\label{sec:implementation_details}

\subsubsection{Learning Strategy}
Our loss function is defined as 
$\textit{L} = \textit{L}_{IoU}^{W} + \textit{L}_{BCE}^{W}$, 
where $\textit{L}_{IoU}^{W}$ and $\textit{L}_{BCE}^{W}$ represent 
the weighted intersection-over-union (IoU) loss and binary cross entropy (BCE) 
loss for the global restriction and local (pixel-level) restriction.
Different from the standard IoU loss, 
which has been widely adopted in segmentation tasks, 
the weighted IoU loss increases the weights of hard pixels 
to highlight their importance.
In addition, compared with the standard BCE loss, 
$\textit{L}_{BCE}^{W}$ pays more attention to hard pixels rather than 
assigning all pixels equal weights. 
The definitions of these losses are the same as in
\cite{qin2019basnet,wei2019f3net} and their effectiveness has been validated 
in the field of salient object detection.
Here, we adopt deep supervision for the three side-outputs 
(\ie, $C_3$, $C_4$, and $C_5$) and the global map $C_6$.
Each map is up-sampled (\eg, $C_3^{up}$) to the same size as the 
ground-truth map $G$.
Thus, the total loss for the proposed \ournewmodel~can be formulated as:
$\textit{L}_{total} = \textit{L} (C_6^{up}, G) + 
\sum_{i=3}^{i=5} \textit{L} (C_i^{up}, G)$.

\subsubsection{Hyperparameter Settings}
\ournewmodel~is implemented in PyTorch and trained with the 
Adam optimizer~\cite{kingma2015adam}.
During the training stage, the batch size is set to 36, 
and the learning rate starts at 1e-4, dividing by 10 every 50 epochs.
The whole training time is only about 4 hours for 100 epochs.
The running time is measured on an Intel$^\circledR$
i9-9820X CPU @3.30GHz $\times$ 20 platform and a single NVIDIA TITAN RTX GPU. 
During inference, each image is resized to 352$\times$352 and then fed into 
the proposed pipeline to obtain the final prediction without 
any post-processing techniques.
The inference speed is $\sim$45 fps on a single GPU without I/O time.
Both PyTorch \cite{paszke2019pytorch} and Jittor \cite{hu2020jittor} 
verisons of the source code will be made publicly avaliable.

\begin{table*}[t!]
  \centering
  \renewcommand{\arraystretch}{0.9}
  \setlength\tabcolsep{9pt}
  \caption{\textbf{Quantitative results on three different datasets.} 
    The best scores are highlighted in \textbf{bold}. 
    Note that the ANet-SRM model (only trained on CAMO) does not have 
    a publicly available code, 
    thus other results are not available. 
    $\uparrow$ indicates the higher the score the better.
    $E_\phi$ denotes mean E-measure~\cite{Fan2018Enhanced}. 
  }\label{tab:ModelScore}
  \vspace{-8pt}
  \begin{tabular}{l|cccc||cccc||cccc} \hline \toprule
    & \multicolumn{4}{c||}{\tabincell{c}{CHAMELEON~\cite{2018Animal}}} 
    & \multicolumn{4}{c||}{\tabincell{c}{CAMO-Test~\cite{le2019anabranch}}} 
    & \multicolumn{4}{c  }{\tabincell{c}{COD10K-Test~(OUR)}} 
    \\ \cline{2-13}
    Baseline Models~~~ 
    & $S_\alpha\uparrow$ &$E_\phi\uparrow$ &$F_\beta^w\uparrow$ &$M\downarrow$
    & $S_\alpha\uparrow$ &$E_\phi\uparrow$ &$F_\beta^w\uparrow$ &$M\downarrow$
    & $S_\alpha\uparrow$ &$E_\phi\uparrow$ &$F_\beta^w\uparrow$ &$M\downarrow$ 
    \\ \hline
    FPN~\cite{lin2017feature}
    &0.794&0.783&0.590&0.075&0.684&0.677&0.483&0.131&0.697&0.691&0.411&0.075\\
    MaskRCNN~\cite{he2017mask}
    &0.643&0.778&0.518&0.099&0.574&0.715&0.430&0.151&0.613&0.748&0.402&0.080\\
    PSPNet~\cite{zhao2017pyramid}
    &0.773&0.758&0.555&0.085&0.663&0.659&0.455&0.139&0.678&0.680&0.377&0.080\\
    UNet++~\cite{zou2018DLMIA}
    &0.695&0.762&0.501&0.094&0.599&0.653&0.392&0.149&0.623&0.672&0.350&0.086\\
    PiCANet~\cite{liu2018picanet}
    &0.769&0.749&0.536&0.085&0.609&0.584&0.356&0.156&0.649&0.643&0.322&0.090\\
    MSRCNN~\cite{huang2019mask}
    &0.637&0.686&0.443&0.091&0.617&0.669&0.454&0.133&0.641&0.706&0.419&0.073\\
    PFANet~\cite{zhao2019pyramid}
    &0.679&0.648&0.378&0.144&0.659&0.622&0.391&0.172&0.636&0.618&0.286&0.128\\
    CPD~\cite{wu2019cascaded}
    &0.853&0.866&0.706&0.052&0.726&0.729&0.550&0.115&0.747&0.770&0.508&0.059\\ 
    HTC~\cite{chen2019hybrid}
    &0.517&0.489&0.204&0.129&0.476&0.442&0.174&0.172&0.548&0.520&0.221&0.088\\
    ANet-SRM~\cite{le2019anabranch}
    & - & - & - & - &0.682&0.685&0.484&0.126& - & - & - & -\\
    EGNet~\cite{zhao2019EGNet}
    &0.848&0.870&0.702&0.050&0.732&0.768&0.583&0.104&0.737&0.779&0.509&0.056\\
    PraNet~\cite{fan2020pranet}     
    &0.860&0.907&0.763&0.044&0.769&0.824&0.663&0.094&0.789&0.861&0.629&0.045\\
    \hline
    \Rev{SINet\_cvpr~\cite{fan2020camouflaged}}     
    & \Rev{0.869} & \Rev{0.891} & \Rev{0.740} & \Rev{0.044} & \Rev{0.751} & \Rev{0.771} & \Rev{0.606} & \Rev{0.100} & \Rev{0.771} & \Rev{0.806} & \Rev{0.551} & \Rev{0.051} \\
    %
    %
    \rowcolor{mygray}\textbf{\ournewmodel~(OUR)}
    & \textbf{0.888} & \textbf{0.942} & \textbf{0.816} & \textbf{0.030}
    & \textbf{0.820} & \textbf{0.882} & \textbf{0.743} & \textbf{0.070}
    & \textbf{0.815} & \textbf{0.887} & \textbf{0.680} & \textbf{0.037} 
    \\ \bottomrule
  \end{tabular}
\end{table*}

\begin{table*}[t!]
  \centering
  \caption{\textbf{Quantitative results on four super-classes of the COD10K 
    dataset in terms of four widely used evaluation metrics.} 
    All methods are trained using the same dataset as in 
    \cite{fan2020camouflaged}. 
    $\uparrow$ indicates the higher the score the better, and 
    $\downarrow$: the lower the better.
  }\label{tab:subclass_cod10k}
  \vspace{-10pt}
  \renewcommand\arraystretch{0.8} 
  \setlength\tabcolsep{4.5pt}
  \begin{tabular}{l|cccc||cccc||cccc||cccc} \hline \toprule
    & \multicolumn{4}{c||}{\tabincell{c}{Amphibian~(124 images)}} 
    & \multicolumn{4}{c||}{\tabincell{c}{Aquatic~(474 images)}}
    & \multicolumn{4}{c||}{\tabincell{c}{Flying~(714 images)}} 
    & \multicolumn{4}{c}{ \tabincell{c}{Terrestrial~(699 images)}}
    \\ \cline{2-17}
    Baseline Models & $S_\alpha\uparrow$ &$E_\phi\uparrow$ &$F_\beta^w\uparrow$ 
    &$M\downarrow$ &$S_\alpha\uparrow$ &$E_\phi\uparrow$ &$F_\beta^w\uparrow$ 
    &$M\downarrow$ &$S_\alpha\uparrow$ &$E_\phi\uparrow$ &$F_\beta^w\uparrow$ 
    &$M\downarrow$ &$S_\alpha\uparrow$ &$E_\phi\uparrow$ &$F_\beta^w\uparrow$ 
    &$M\downarrow$\\
    \hline
    FPN~\cite{lin2017feature} 
    & 0.745 & 0.776 & 0.497 & 0.065 & 0.684 & 0.732 & 0.432 & 0.103 
    & 0.726 & 0.766 & 0.440 & 0.061 & 0.601 & 0.656 & 0.353 & 0.109\\
    MaskRCNN~\cite{he2017mask} 
    & 0.665 & 0.785 & 0.487 & 0.081 & 0.560 & 0.721 & 0.344 & 0.123 
    & 0.644 & 0.767 & 0.449 & 0.063 & 0.611 & 0.630 & 0.380 & 0.075\\
    PSPNet~\cite{zhao2017pyramid} 
    & 0.736 & 0.774 & 0.463 & 0.072 & 0.659 & 0.712 & 0.396 & 0.111 
    & 0.700 & 0.743 & 0.394 & 0.067 & 0.669 & 0.718 & 0.332 & 0.071\\
    UNet++~\cite{zou2018DLMIA} 
    & 0.677 & 0.745 & 0.434 & 0.079 & 0.599 & 0.673 & 0.347 & 0.121 
    & 0.659 & 0.727 & 0.397 & 0.068 & 0.608 & 0.749 & 0.288 & 0.070\\
    PiCANet~\cite{liu2018picanet} 
    & 0.686 & 0.702 & 0.405 & 0.079 & 0.616 & 0.631 & 0.335 & 0.115 
    & 0.663 & 0.676 & 0.347 & 0.069 & 0.658 & 0.708 & 0.273 & 0.074\\
    MSRCNN~\cite{huang2019mask} 
    & 0.722 & 0.786 & 0.555 & 0.055 & 0.614 & 0.686 & 0.398 & 0.107 
    & 0.675 & 0.744 & 0.466 & 0.058 & 0.594 & 0.661 & 0.361 & 0.081\\
    PFANet~\cite{zhao2019pyramid} 
    & 0.693 & 0.677 & 0.358 & 0.110 & 0.629 & 0.626 & 0.319 & 0.155 
    & 0.658 & 0.648 & 0.299 & 0.102 & 0.611 & 0.603 & 0.237 & 0.111\\
    CPD~\cite{wu2019cascaded} 
    & 0.794 & 0.839 & 0.587 & 0.051 & 0.739 & 0.792 & 0.529 & 0.082 
    & 0.777 & 0.827 & 0.544 & 0.046 & 0.714 & 0.771 & 0.445 & 0.058\\
    HTC~\cite{chen2019hybrid} 
    & 0.606 & 0.598 & 0.331 & 0.088 & 0.507 & 0.495 & 0.183 & 0.129 
    & 0.582 & 0.559 & 0.274 & 0.070 & 0.530 & 0.485 & 0.170 & 0.078\\
    EGNet~\cite{zhao2019EGNet} 
    & 0.785 & 0.854 & 0.606 & 0.047 & 0.725 & 0.793 & 0.528 & 0.080 
    & 0.766 & 0.826 & 0.543 & 0.044 & 0.700 & 0.775 & 0.445 & 0.053\\
    PraNet~\cite{fan2020pranet} 
    & 0.842 & 0.905 & 0.717 & 0.035 & 0.781 & 0.883 & 0.696 & 0.065 
    & 0.819 & 0.888 & 0.669 & 0.033 & 0.756 & 0.835 & 0.565 & 0.046\\
    \hline
    \Rev{SINet\_cvpr~\cite{fan2020camouflaged}}     
    & \Rev{0.827} & \Rev{0.866} & \Rev{0.654} & \Rev{0.042} & \Rev{0.758} & \Rev{0.803} & \Rev{0.570} & \Rev{0.073} & \Rev{0.798} & \Rev{0.828} & \Rev{0.580} & \Rev{0.040} & \Rev{0.743} & \Rev{0.778} & \Rev{0.491} & \Rev{0.050} \\
    \rowcolor{mygray}
    \textbf{\ournewmodel~(OUR)} 
    & \textbf{0.858} & \textbf{0.916} & \textbf{0.756} & \textbf{0.030} 
    & \textbf{0.811} & \textbf{0.883} & \textbf{0.696} & \textbf{0.051} 
    & \textbf{0.839} & \textbf{0.908} & \textbf{0.713} & \textbf{0.027} 
    & \textbf{0.787} & \textbf{0.866} & \textbf{0.623} & \textbf{0.039}
    \\ \bottomrule
  \end{tabular}
\end{table*}

\section{COD Benchmark}\label{sec:benchmark}

\subsection{Experimental Settings}\label{sec:benchmarkSetup}

\subsubsection{Evaluation Metrics}

Mean absolute error (MAE) is widely used in SOD tasks.
Following Perazzi~\etal~\cite{perazzi2012saliency}, 
we also adopt the MAE ($M$) metric to assess the pixel-level accuracy 
between a predicted map and ground-truth.
However, while useful for assessing the presence and amount of error, 
the MAE metric is not able to determine where the error occurs.
Recently, Fan~\etal proposed a human visual perception based 
E-measure ($E_\phi$)~\cite{Fan2018Enhanced}, 
which simultaneously evaluates the pixel-level matching and 
image-level statistics.
This metric is naturally suited for assessing the overall and
localized accuracy of the concealed object detection results.
Note that we report mean $E_\phi$ in the experiments.
Since concealed objects often contain complex shapes, 
COD also requires a metric that can judge structural similarity.
We therefore utilize the S-measure ($S_\alpha$)~\cite{fan2017structure}
as our structural similarity evaluation metric.
Finally, recent studies~\cite{Fan2018Enhanced,fan2017structure} have suggested 
that the weighted F-measure ($F_\beta^w$)~\cite{margolin2014evaluate} 
can provide more reliable evaluation results than the traditional $F_\beta$.
Thus, we further consider this as an alternative metric for COD.
Our one-key evaluation code is also available at the project page.

\subsubsection{Baseline Models}

We select \baselineN~deep learning baselines~\cite{lin2017feature,he2017mask,
zhao2017pyramid,zou2018DLMIA,liu2018picanet,huang2019mask,zhao2019pyramid,
wu2019cascaded,chen2019hybrid,zhao2019EGNet,le2019anabranch,fan2020pranet} according to the following criteria:
\textit{a)} classical architectures,
\textit{b)} recently published, and
\textit{c)} achieve SOTA performance in a specific field.

\subsubsection{Training/Testing Protocols}
For fair comparison with our previous version~\cite{fan2020camouflaged},
we adopt the same training settings~\cite{fan2020camouflaged} for the baselines.\footnote{To verify the generalizability of \ournewmodel, we only use the combined training set of CAMO~\cite{le2019anabranch} and 
COD10K~\cite{fan2020camouflaged} without EXTRA (\ie, additional) data.}
We evaluate the models on the whole CHAMELEON~\cite{2018Animal} dataset 
and the test sets of CAMO and COD10K.
%
%

\begin{figure*}[t!]
  \centering
  \includegraphics[width=\linewidth]{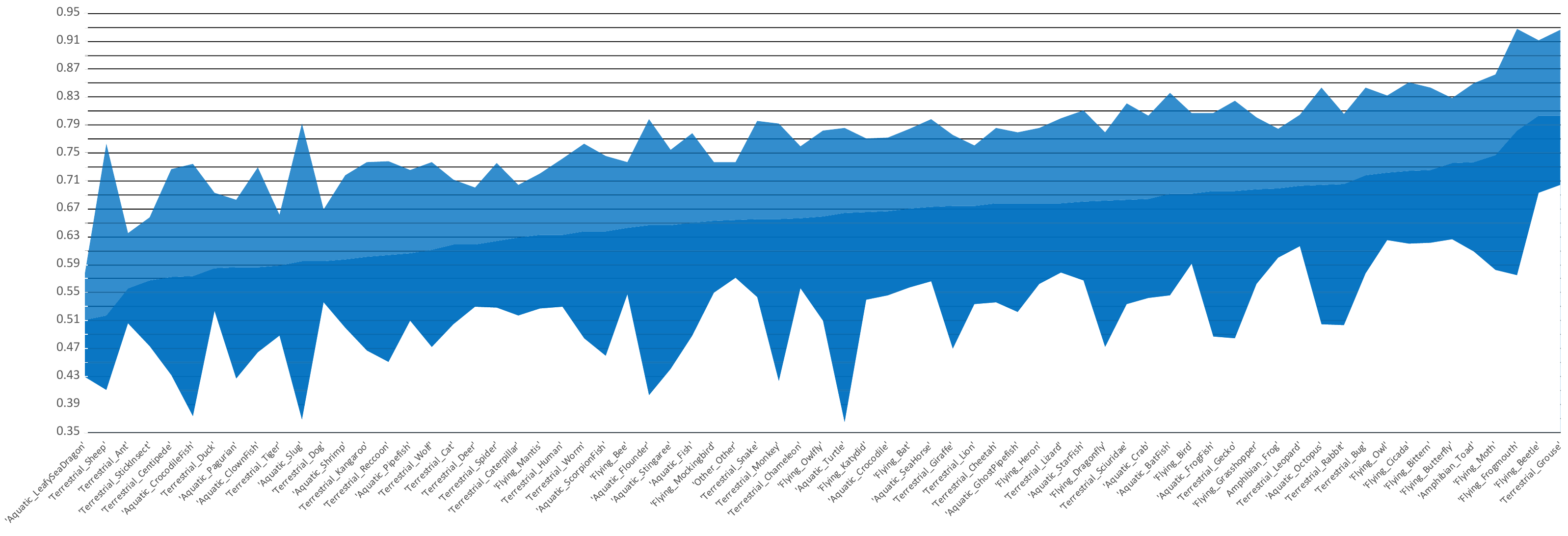}
  \vspace{-15pt}
  \caption{\textbf{Per-subclass performance.}
    Sub-classes are sorted by \emph{difficulty}, 
    determined by the mean $S_\alpha$~\cite{fan2017structure} across 
    \baselineN~baselines. 
    We also provide the minimum (bottom line) and maximum (top line) 
    $S_\alpha$ for each sub-class.
  }\label{fig:subClassPerformance}
\end{figure*}

\subsection{Results and Data Analysis}
This section provides the quantitative evaluation results on CHAMELEON, 
CAMO, and COD10K datasets, respectively.

\myPara{Performance on CHAMELEON.}
From \tabref{tab:ModelScore}, compared with the \baselineN~SOTA object 
detection baselines and ANet-SRM, 
our \ournewmodel~achieves the new SOTA performances across all metrics.
Note that our model does not apply any auxiliary edge/boundary features 
(\eg, EGNet~\cite{zhao2019EGNet}, PFANet~\cite{zhao2019pyramid}), 
pre-processing techniques~\cite{mori2005guiding}, 
or post-processing strategies such as~\cite{kra2011efficient,boykov1999fast}.


\begin{table*}[thp!]
  \centering
  \caption{\textbf{Results of $S_\alpha$ for each sub-class in our 
    \emph{COD10K} dataset.} 
    The best performing method of each category is highlighted in \textbf{bold}.
  }\label{tab:Atr_eachimage1}
  \renewcommand{\arraystretch}{1.05}
  \renewcommand{\tabcolsep}{3.1pt}
  \vspace{-8pt}
  \begin{tabular}{l | ccccc ccccc cccc}
  \toprule
    &HTC & PFANet & MRCNN & BASNet &UNet++
    &PiCANet &MSRCNN &PoolNet &PSPNet &FPN
    &EGNet &CPD &PraNet & \ournewmodel\\
    \specialrule{0em}{-0.5pt}{-2pt}
    \textbf{\emph{Sub-class}} 
    &\cite{chen2019hybrid} &\cite{zhao2019pyramid} &\cite{he2017mask} &\cite{qin2019basnet} 
    &\cite{zou2018DLMIA} &\cite{liu2018picanet} &\cite{huang2019mask} &\cite{liu2019simple} &\cite{zhao2017pyramid} 
    &\cite{lin2017feature} &\cite{zhao2019EGNet} &\cite{wu2019cascaded} &\cite{fan2020pranet} &\textbf{OUR}\\
   \hline
Amphibian-Frog &0.600 &0.678 &0.664 &0.692 &0.656 &0.687 &0.692 &0.732 &0.697 &0.731 &0.745 &0.752 & \Rev{\textit{0.823}} &\textbf{0.837}\\
\rowcolor{mygray}
Amphibian-Toad &0.609 &0.697 &0.666 &0.717 &0.689 &0.714 &0.739 &0.786 &0.757 &0.752 &0.812 &0.817 & \Rev{\textit{0.853}} &\textbf{0.870}\\
Aquatic-BatFish &0.546 &0.746 &0.634 &0.749 &0.626 &0.624 &0.637 &0.741 &0.724 &0.764 &0.707 &0.761 &\textbf{0.879} & \Rev{\textit{0.873}}\\
\rowcolor{mygray}
Aquatic-ClownFish &0.547 &0.519 &0.509 &0.464 &0.548 &0.636 &0.571 &0.626 &0.531 &0.730 &0.632 &0.646 & \Rev{\textit{0.707}} &\textbf{0.787}\\
Aquatic-Crab &0.543 &0.661 &0.691 &0.643 &0.630 &0.675 &0.634 &0.727 &0.680 &0.724 &0.760 &0.753 & \Rev{\textit{0.792}} &\textbf{0.815}\\
\rowcolor{mygray}
Aquatic-Crocodile &0.546 &0.631 &0.660 &0.599 &0.602 &0.669 &0.646 &0.743 &0.636 &0.687 &0.772 &0.761 & \Rev{\textit{0.806}} &\textbf{0.825}\\
Aquatic-CrocodileFish &0.436 &0.572 &0.558 &0.475 &0.559 &0.373 &0.479 &0.693 &0.624 &0.515 & \Rev{\textit{0.709}} &0.690 &0.669 &\textbf{0.746}\\
\rowcolor{mygray}
Aquatic-Fish &0.488 &0.622 &0.597 &0.625 &0.574 &0.619 &0.680 &0.703 &0.650 &0.699 &0.717 &0.778 & \Rev{\textit{0.784}} &\textbf{0.834}\\
Aquatic-Flounder &0.403 &0.663 &0.539 &0.633 &0.569 &0.704 &0.570 &0.782 &0.646 &0.695 &0.798 &0.774 & \Rev{\textit{0.835}} &\textbf{0.889}\\
\rowcolor{mygray}
Aquatic-FrogFish &0.595 &0.768 &0.650 &0.736 &0.671 &0.670 &0.653 &0.719 &0.695 & \Rev{\textit{0.807}} &0.806 &0.730 &0.781 &\textbf{0.894}\\
Aquatic-GhostPipefish &0.522 &0.690 &0.556 &0.679 &0.651 &0.675 &0.636 &0.717 &0.709 &0.744 &0.759 &0.763 & \Rev{\textit{0.784}} &\textbf{0.817}\\
\rowcolor{mygray}
Aquatic-LeafySeaDragon &0.460 &0.576 &0.442 &0.481 &0.523 &0.499 &0.500 &0.547 &0.563 &0.507 &0.522 &0.534 & \Rev{\textit{0.587}} &\textbf{0.670}\\
Aquatic-Octopus &0.505 &0.708 &0.598 &0.644 &0.663 &0.673 &0.720 &0.779 &0.723 &0.760 &0.810 &0.812 &\textbf{0.896} & \Rev{\textit{0.887}}\\
\rowcolor{mygray}
Aquatic-Pagurian &0.427 &0.578 &0.477 &0.607 &0.553 &0.624 &0.583 &0.657 &0.608 &0.638 & \Rev{\textit{0.683}} &0.611 &0.615 &\textbf{0.698}\\
Aquatic-Pipefish &0.510 &0.553 &0.531 &0.557 &0.550 &0.612 &0.566 &0.625 &0.642 &0.632 &0.681 &0.704 & \Rev{\textit{0.769}} &\textbf{0.781}\\
\rowcolor{mygray}
Aquatic-ScorpionFish &0.459 &0.697 &0.482 &0.686 &0.630 &0.605 &0.600 &0.729 &0.649 &0.668 &0.730 &0.746 & \Rev{\textit{0.766}} &\textbf{0.808}\\
Aquatic-SeaHorse &0.566 &0.656 &0.581 &0.664 &0.663 &0.623 &0.657 &0.698 &0.687 &0.715 &0.750 &0.765 & \Rev{\textit{0.810}} &\textbf{0.823}\\
\rowcolor{mygray}
Aquatic-Shrimp &0.500 &0.574 &0.520 &0.631 &0.586 &0.574 &0.546 &0.605 &0.591 &0.667 &0.647 &0.669 & \Rev{\textit{0.727}} &\textbf{0.735}\\
Aquatic-Slug &0.493 &0.581 &0.368 &0.492 &0.533 &0.460 &0.661 &0.732 &0.547 &0.664 &\textbf{0.777} & \Rev{\textit{0.774}} &0.701 &0.729\\
\rowcolor{mygray}
Aquatic-StarFish &0.568 &0.641 &0.617 &0.611 &0.657 &0.638 &0.580 &0.733 &0.722 &0.756 & \Rev{\textit{0.811}} &0.787 &0.779 &\textbf{0.890}\\
Aquatic-Stingaree &0.519 &0.721 &0.670 &0.618 &0.571 &0.569 &0.709 &0.733 &0.616 &0.670 &0.741 & \Rev{\textit{0.754}} &0.704 &\textbf{0.815}\\
\rowcolor{mygray}
Aquatic-Turtle &0.364 &0.686 &0.594 &0.658 &0.565 &0.734 &0.762 &0.757 &0.664 &0.745 &0.752 &\textbf{0.786} & \Rev{\textit{0.773}} &0.760\\
Flying-Bat &0.589 &0.652 &0.611 &0.623 &0.557 &0.638 &0.679 &0.725 &0.657 &0.714 &0.765 &0.784 & \Rev{\textit{0.817}} &\textbf{0.847}\\
\rowcolor{mygray}
Flying-Bee &0.578 &0.579 &0.628 &0.547 &0.588 &0.616 &0.679 &0.670 &0.655 &0.665 &0.737 &0.709 & \Rev{\textit{0.763}} &\textbf{0.777}\\
Flying-Beetle &0.699 &0.741 &0.693 &0.810 &0.829 &0.780 &0.796 &0.860 &0.808 &0.848 &0.830 &0.887 & \Rev{\textit{0.890}} &\textbf{0.903}\\
\rowcolor{mygray}
Flying-Bird &0.591 &0.628 &0.680 &0.627 &0.643 &0.674 &0.681 &0.735 &0.696 &0.708 &0.763 &0.785 & \Rev{\textit{0.822}} &\textbf{0.835}\\
Flying-Bittern &0.639 &0.621 &0.703 &0.650 &0.673 &0.741 &0.704 &0.785 &0.701 &0.751 &0.802 & \Rev{\textit{0.838}} &0.827 &\textbf{0.849}\\
\rowcolor{mygray}
Flying-Butterfly &0.653 &0.692 &0.697 &0.700 &0.725 &0.714 &0.762 &0.777 &0.736 &0.758 &0.816 &0.818 & \Rev{\textit{0.871}} &\textbf{0.883}\\
Flying-Cicada &0.640 &0.682 &0.620 &0.729 &0.675 &0.691 &0.708 &0.781 &0.744 &0.733 &0.820 &0.812 & \Rev{\textit{0.845}} &\textbf{0.883}\\
\rowcolor{mygray}
Flying-Dragonfly &0.472 &0.679 &0.624 &0.712 &0.670 &0.694 &0.682 &0.695 &0.681 &0.707 &0.761 & \Rev{\textit{0.779}} & \Rev{\textit{0.779}} &\textbf{0.837}\\
Flying-Frogmouth &0.684 &0.766 &0.648 &0.828 &0.813 &0.722 &0.773 &0.883 &0.741 &0.795 &0.901 & \Rev{\textit{0.928}} &0.927 &\textbf{0.941}\\
\rowcolor{mygray}
Flying-Grasshopper &0.563 &0.671 &0.651 &0.689 &0.656 &0.692 &0.666 &0.734 &0.710 &0.740 &0.773 &0.779 & \Rev{\textit{0.821}} &\textbf{0.833}\\
Flying-Heron &0.563 &0.579 &0.629 &0.598 &0.670 &0.647 &0.699 &0.718 &0.654 &0.743 &0.783 &0.786 & \Rev{\textit{0.810}} &\textbf{0.823}\\
\rowcolor{mygray}
Flying-Katydid &0.540 &0.661 &0.593 &0.657 &0.653 &0.659 &0.615 &0.696 &0.687 &0.709 &0.730 &0.739 & \Rev{\textit{0.802}} &\textbf{0.809}\\
Flying-Mantis &0.527 &0.622 &0.569 &0.618 &0.614 &0.629 &0.603 &0.661 &0.658 &0.670 &0.696 &0.690 & \Rev{\textit{0.749}} &\textbf{0.775}\\
\rowcolor{mygray}
Flying-Mockingbird &0.641 &0.550 &0.622 &0.593 &0.636 &0.596 &0.664 &0.670 &0.674 &0.683 &0.721 &0.737 & \Rev{\textit{0.788}} &\textbf{0.838}\\
Flying-Moth &0.583 &0.720 &0.726 &0.737 &0.707 &0.685 &0.747 &0.783 &0.753 &0.798 &0.833 &0.854 & \Rev{\textit{0.878}} &\textbf{0.917}\\
\rowcolor{mygray}
Flying-Owl &0.625 &0.671 &0.705 &0.656 &0.657 &0.718 &0.710 &0.781 &0.712 &0.750 &0.793 &0.809 & \Rev{\textit{0.837}} &\textbf{0.868}\\
Flying-Owlfly &0.614 &0.690 &0.524 &0.669 &0.633 &0.580 &0.599 &0.778 &0.583 &0.743 & \Rev{\textit{0.782}} &0.756 &0.758 &\textbf{0.863}\\
\rowcolor{mygray}
Other-Other &0.571 &0.613 &0.603 &0.593 &0.638 &0.653 &0.675 &0.687 &0.671 &0.665 &0.725 &0.700 & \Rev{\textit{0.777}} &\textbf{0.779}\\
Terrestrial-Ant &0.506 &0.516 &0.508 &0.519 &0.523 &0.585 &0.538 &0.552 &0.572 &0.564 &0.627 &0.605 &\textbf{0.676} & \Rev{\textit{0.669}}\\
\rowcolor{mygray}
Terrestrial-Bug &0.578 &0.681 &0.682 &0.687 &0.686 &0.701 &0.691 &0.743 &0.710 &0.799 &0.799 &0.803 & \Rev{\textit{0.828}} &\textbf{0.856}\\
Terrestrial-Cat &0.505 &0.585 &0.591 &0.557 &0.562 &0.608 &0.613 &0.669 &0.624 &0.634 &0.682 &0.678 & \Rev{\textit{0.745}} &\textbf{0.772}\\
\rowcolor{mygray}
Terrestrial-Caterpillar &0.517 &0.643 &0.569 &0.691 &0.636 &0.581 &0.575 &0.638 &0.640 &0.685 &0.684 &0.704 & \Rev{\textit{0.729}} &\textbf{0.776}\\
Terrestrial-Centipede &0.432 &0.573 &0.476 &0.485 &0.496 &0.554 &0.629 &0.703 &0.561 &0.536 &0.727 &0.643 & \Rev{\textit{0.704}} &\textbf{0.762}\\
\rowcolor{mygray}
Terrestrial-Chameleon &0.556 &0.651 &0.627 &0.653 &0.619 &0.619 &0.632 &0.695 &0.659 &0.673 &0.713 &0.732 & \Rev{\textit{0.789}} &\textbf{0.804}\\
Terrestrial-Cheetah &0.536 &0.649 &0.699 &0.624 &0.603 &0.662 &0.598 &0.717 &0.720 &0.667 &0.732 &0.769 & \Rev{\textit{0.800}} &\textbf{0.826}\\
\rowcolor{mygray}
Terrestrial-Deer &0.530 &0.581 &0.610 &0.564 &0.558 &0.600 &0.623 &0.650 &0.644 &0.660 &0.667 &0.670 & \Rev{\textit{0.719}} &\textbf{0.757}\\
Terrestrial-Dog &0.572 &0.560 &0.596 &0.536 &0.559 &0.574 &0.614 &0.608 &0.588 &0.613 &0.607 &0.648 & \Rev{\textit{0.666}} &\textbf{0.707}\\
\rowcolor{mygray}
Terrestrial-Duck &0.530 &0.535 &0.557 &0.539 &0.524 &0.558 &0.619 &0.582 &0.602 &0.548 &0.598 &0.682 & \Rev{\textit{0.742}} &\textbf{0.746}\\
Terrestrial-Gecko &0.485 &0.674 &0.662 &0.725 &0.683 &0.705 &0.606 &0.733 &0.724 &0.747 &0.789 &0.771 & \Rev{\textit{0.833}} &\textbf{0.848}\\
\rowcolor{mygray}
Terrestrial-Giraffe &0.469 &0.628 &0.697 &0.620 &0.611 &0.701 &0.635 &0.681 &0.718 &0.722 &0.747 &0.776 &\textbf{0.809} & \Rev{\textit{0.784}}\\
Terrestrial-Grouse &0.704 &0.760 &0.726 &0.721 &0.774 &0.805 &0.780 &0.879 &0.803 &0.806 &0.904 & \Rev{\textit{0.919}} &0.888 &\textbf{0.921}\\
\rowcolor{mygray}
Terrestrial-Human &0.530 &0.629 &0.608 &0.613 &0.549 &0.577 &0.658 &0.697 &0.636 &0.665 &0.708 &0.700 & \Rev{\textit{0.765}} &\textbf{0.817}\\
Terrestrial-Kangaroo &0.482 &0.586 &0.599 &0.467 &0.548 &0.588 &0.571 &0.644 &0.630 &0.623 &0.650 &0.620 & \Rev{\textit{0.798}} &\textbf{0.816}\\
\rowcolor{mygray}
Terrestrial-Leopard &0.617 &0.647 &0.742 &0.616 &0.640 &0.652 &0.673 &0.736 &0.720 &0.704 &0.744 & \Rev{\textit{0.791}} & \Rev{\textit{0.791}} &\textbf{0.823}\\
Terrestrial-Lion &0.534 &0.634 &0.695 &0.599 &0.660 &0.656 &0.658 &0.720 &0.714 &0.663 &0.754 &0.751 & \Rev{\textit{0.805}} &\textbf{0.813}\\
\rowcolor{mygray}
Terrestrial-Lizard &0.579 &0.629 &0.634 &0.635 &0.633 &0.656 &0.627 &0.710 &0.702 &0.716 &0.744 &0.777 & \Rev{\textit{0.804}} &\textbf{0.830}\\
Terrestrial-Monkey &0.423 &0.693 &0.724 &0.593 &0.611 &0.730 &0.663 &0.792 &0.678 &0.614 &0.709 &0.699 & \Rev{\textit{0.851}} &\textbf{0.888}\\
\rowcolor{mygray}
Terrestrial-Rabbit &0.504 &0.657 &0.685 &0.634 &0.635 &0.721 &0.731 &0.794 &0.722 &0.758 &0.789 &0.806 & \Rev{\textit{0.829}} &\textbf{0.843}\\
Terrestrial-Reccoon &0.451 &0.525 &0.536 &0.461 &0.482 &0.702 &0.723 &0.643 &0.532 &0.592 &0.691 &0.659 &\textbf{0.781} & \Rev{\textit{0.766}}\\
\rowcolor{mygray}
Terrestrial-Sciuridae &0.533 &0.612 &0.638 &0.573 &0.608 &0.693 &0.661 &0.745 &0.725 &0.721 &0.775 &0.757 & \Rev{\textit{0.810}} &\textbf{0.842}\\
Terrestrial-Sheep &0.434 &0.451 &0.721 &0.410 &0.482 &0.467 &\textbf{0.763} & \Rev{\textit{0.660}} &0.466 &0.430 &0.489 &0.487 &0.481 &0.500\\
\rowcolor{mygray}
Terrestrial-Snake &0.544 &0.590 &0.586 &0.603 &0.567 &0.614 &0.597 &0.714 &0.695 &0.652 &0.738 & \Rev{\textit{0.788}} &0.771 &\textbf{0.831}\\
Terrestrial-Spider &0.528 &0.594 &0.593 &0.594 &0.580 &0.621 &0.572 &0.650 &0.649 &0.651 &0.685 &0.687 & \Rev{\textit{0.740}} &\textbf{0.771}\\
\rowcolor{mygray}
Terrestrial-StickInsect &0.473 &0.548 &0.486 &0.526 &0.535 &0.600 &0.491 &0.578 &0.607 &0.629 &0.616 &0.647 & \Rev{\textit{0.660}} &\textbf{0.696}\\
Terrestrial-Tiger &0.489 &0.583 &0.576 &0.555 &0.573 &0.563 &0.565 &0.638 &0.602 &0.599 &0.647 &0.621 & \Rev{\textit{0.690}} &\textbf{0.703}\\
\rowcolor{mygray}
Terrestrial-Wolf &0.472 &0.574 &0.602 &0.535 &0.534 &0.568 &0.621 &0.650 &0.656 &0.651 &0.704 &0.662 & \Rev{\textit{0.737}} &\textbf{0.749}\\
Terrestrial-Worm &0.485 &0.652 &0.596 &0.642 &0.628 &0.558 &0.651 &0.692 &0.629 &0.684 &0.763 &0.670 & \Rev{\textit{0.724}} &\textbf{0.806}\\

  \bottomrule
  \end{tabular}
\end{table*}

\myPara{Performance on CAMO.}
We also test our model on the CAMO~\cite{le2019anabranch} dataset, 
which includes various concealed objects.
Based on the overall performances reported in \tabref{tab:ModelScore}, 
we find that the CAMO dataset is more challenging than CHAMELEON.
Again, \ournewmodel~obtains the best performance, 
further demonstrating its robustness.

\myPara{Performance on COD10K.}
With the test set (2,026 images) of our COD10K dataset, 
we again observe that the proposed \ournewmodel~is consistently 
better than other competitors.
This is because its specially designed search and identification modules 
can automatically learn rich diversified features from coarse to fine, 
which are crucial for overcoming challenging ambiguities in object boundaries.
The results are shown in \tabref{tab:ModelScore} and 
\tabref{tab:subclass_cod10k}.

\myPara{Per-subclass Performance.}
In addition to the overall quantitative comparisons on our COD10K dataset, 
we also report the quantitative per-subclass results in the 
\tabref{tab:Atr_eachimage1} to investigate the pros and cons of the models 
for future researchers.
In \figref{fig:subClassPerformance}, we additionally show the minimum, mean, 
and maximum S-measure performance of each sub-class over all baselines.
The \Rev{easiest} sub-class is ``Grouse'', 
while the most difficult is the ``LeafySeaDragon'', 
from the aquatic and terrestrial categories, respectively.

\myPara{Qualitative Results.}
We present more detection results of our conference version model (SINet\_cvpr) 
for various challenging concealed objects, such as \emph{spider}, \emph{moth}, 
\emph{sea horse}, and \emph{toad}, in the \supp{supplementary materials}.
As shown in \figref{fig:ComparedSOTA}, \ournewmodel~further improves the visual 
results compared to SINet\_cvpr~in terms of different lighting ($1^{st}$ row), 
appearance changes ($2^{nd}$ row), 
and indefinable boundaries ($3^{rd}$ to $5^{th}$).  
PFANet~\cite{zhao2019pyramid} is able to locate the concealed objects, 
but the outputs are always inaccurate.
By further using reverse attention module, PraNet~\cite{fan2020pranet} 
achieves a relatively more accurate location than PFANet in the first case. 
Nevertheless, it still misses the fine details of objects,
especially for the \emph{fish} in the $2^{nd}$ and $3^{rd}$ rows.
For all these challenging cases, \ournewmodel~is able to
infer the real concealed object with fine details, 
demonstrating the robustness of our framework.

\myPara{GOS \textit{vs.} SOD Baselines.}
One noteworthy finding is that, among the top-3 models, 
the GOS model (\ie, FPN~\cite{lin2017feature})
performs worse than the SOD competitors, CPD~\cite{wu2019cascaded}, 
EGNet~\cite{zhao2019EGNet},
suggesting that the SOD framework may be better suited for extension to COD tasks.
Compared with both the GOS~\cite{lin2017feature,he2017mask,zhao2017pyramid,
zou2018DLMIA,huang2019mask,chen2019hybrid} 
and the SOD~\cite{liu2018picanet,zhao2019pyramid,wu2019cascaded,zhao2019EGNet}
models, \ournewmodel~significantly decrease the training time
(\eg, \ournewmodel: 4 hours \textit{vs.} EGNet: 48 hours) and 
achieve the SOTA performance on all datasets, 
showing that they are promising solutions for the COD problem. 
Due to the limited space, fully comparing them with existing SOTA SOD models 
is beyond the scope of this paper. 
Note that our main goal is to provide more general observations for 
future work. 
More recent SOD models can be found in our project page.

\begin{figure*}[t!]
  \centering
  \includegraphics[width=.99\textwidth]{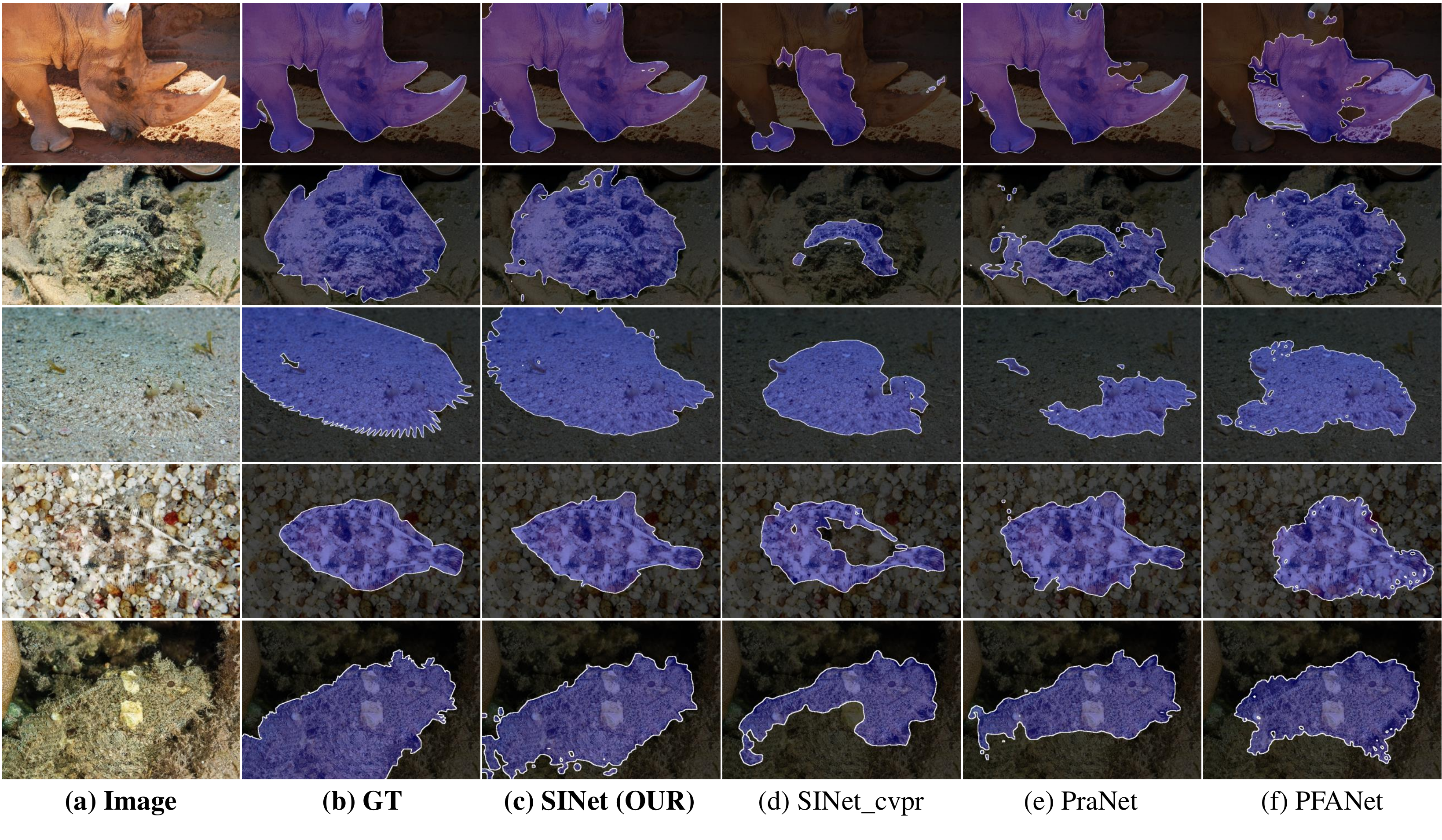}
  \vspace*{-15pt}
  \caption{\textbf{Comparison of our \ournewmodel~and 
    three top-performing baselines}, 
    including (d) SINet\_cvpr~\cite{fan2020camouflaged}, 
    (e) PraNet~\cite{fan2020pranet}, and (f) PFANet~\cite{zhao2019pyramid}.
  }\label{fig:ComparedSOTA}
\end{figure*}

\begin{table}[t!]
  \centering
  \renewcommand{\arraystretch}{0.9}
  \setlength\tabcolsep{4.5pt}
  \caption{\textbf{Structure-measure 
    ($S_\alpha\uparrow$~\cite{fan2017structure}) scores for cross-dataset 
    generalization.} 
    SINet\_cvpr~is trained on one dataset (rows) and tested on all datasets 
    (columns). 
    ``Self'': training and testing on the same dataset (diagonal). 
    ``Mean others'': average score on all except self.
  }\label{tab:DatasetGeneralization}
  \vspace{-8pt}
  \begin{tabular}{r|cc|cc|c} \hline \toprule
    \diagbox{Trained on:}{Tested on:} 
    & \tabincell{c}{CAMO\\~\cite{le2019anabranch}}
    & \tabincell{c}{COD10K\\ (OUR)} & Self 
    & \tabincell{c}{Mean\\others} & Drop$\downarrow$\\ \hline
    CAMO~\cite{le2019anabranch} &\underline{0.803}&0.702&0.803&\Rev{0.702}&\Rev{12.6\%}\\
    \ourdataset~(OUR) &0.742&\underline{0.700}&0.700&\Rev{0.742}&\Rev{-6.0\%}\\
    \hline
    Mean \Rev{others} &\Rev{0.742}&\Rev{0.702}&  & &\\
  \bottomrule
  \end{tabular}
\end{table}

\myPara{Generalization.}
The generalizability and difficulty of datasets play a crucial role in 
both training and assessing different algorithms~\cite{wang2019salient}.
Hence, we study these aspects for existing COD datasets, 
using the cross-dataset analysis method~\cite{torralba2011unbiased}, \ie, 
training a model on one dataset, and testing it on others.
We select two datasets, namely CAMO~\cite{le2019anabranch}, and our \ourdataset.
Following~\cite{wang2019salient}, for each dataset, we randomly
select 800 images as the training set and 200 images as the testing set.
For fair comparison, we train SINet\_cvpr on each dataset
until the loss is stable.

\begin{table*}[t!]
  \centering
  \renewcommand{\arraystretch}{0.9}
  \setlength\tabcolsep{3.1pt}
  \caption{\textbf{Ablation studies for each component on three test datasets.} 
    For details please refer to~\secref{sec:ablation_study}.
  }\label{tab:ablation}
  \vspace{-8pt}
  \begin{tabular}{l|cc|cc|cc||cccc|cccc|cccc} \toprule
    & \multicolumn{2}{c|}{\tabincell{c}{Decoder}} 
    & \multicolumn{2}{c|}{\tabincell{c}{TEM}} 
    & \multicolumn{2}{c||}{\tabincell{c}{GRA}} 
    & \multicolumn{4}{c|}{\tabincell{c}{CHAMELEON~\cite{2018Animal}}} 
    & \multicolumn{4}{c|}{\tabincell{c}{CAMO-Test~\cite{le2019anabranch}}}  
    & \multicolumn{4}{c}{\tabincell{c}{COD10K-Test (OUR)}}\\
    \cline{2-19}
    No. & PD & NCD & Sy. Conv. & Asy. Conv, & Reverse & Group Size
    &$S_\alpha\uparrow$ &$E_\phi\uparrow$ &$F_\beta^w\uparrow$ &$M\downarrow$
    &$S_\alpha\uparrow$ &$E_\phi\uparrow$ &$F_\beta^w\uparrow$ &$M\downarrow$
    &$S_\alpha\uparrow$ &$E_\phi\uparrow$ &$F_\beta^w\uparrow$ &$M\downarrow$\\
    \hline
    \#1 &  & & & $\checkmark$ & $\{ 1,0,0 \}$ & $\{ 32;8;1 \}$
    & 0.884   & 0.940   & 0.811   & 0.033
    & 0.812   & 0.869   & 0.730   & 0.073
    & 0.812   & 0.884   & 0.679   & 0.039\\ 
    \#2 & $\checkmark$ & & & $\checkmark$ & $\{ 1,0,0 \}$ & $\{ 32;8;1 \}$
    & 0.881   & 0.934   & 0.799   & 0.034
    & \Rev{\textbf{0.820}}   & 0.877   & 0.740   & 0.071
    & 0.813   & 0.884   & 0.673   & 0.038\\
    \hline
    \#3 & & $\checkmark$ & & & $\{ 1,0,0 \}$ & $\{ 32;8;1 \}$
    & 0.887   & 0.934   & 0.813   & 0.033
    & 0.811   & 0.867   & 0.731   & 0.074
    & \Rev{\textbf{0.815}}   & \Rev{\textbf{0.888}}   & 0.680  & \Rev{\textbf{0.036}}\\
    \#4 & & $\checkmark$ & $\checkmark$ & & $\{ 1,0,0 \}$ & $\{ 32;8;1 \}$
    & \Rev{\textbf{0.888}}   & \Rev{\textbf{0.944}}   & \Rev{\textbf{0.818}}   & \Rev{\textbf{0.030}}
    & 0.810   & 0.866   & 0.730   & 0.073
    & 0.814   & 0.883   & 0.678   & 0.037\\
    \hline
    \#5 & & $\checkmark$ & & $\checkmark$ & $\{ 0,0,0 \}$ & $\{ 32;8;1 \}$
    & 0.886   & 0.942   & 0.814   & 0.031
    & 0.814   & 0.873   & 0.739   & 0.073
    & 0.814   & 0.887   & \Rev{\textbf{0.682}}   & 0.037\\
    \#6 & & $\checkmark$ & & $\checkmark$ & $\{ 1,1,0 \}$ & $\{ 32;8;1 \}$
    & 0.879   & 0.928   & 0.794   & 0.035
    & \Rev{\textbf{0.820}}   & 0.877   & 0.738   & 0.071
    & 0.807   & 0.878   & 0.661   & 0.040\\
    \#7 & & $\checkmark$ & & $\checkmark$ & $\{ 1,1,1 \}$ & $\{ 32;8;1 \}$
    & 0.886   & 0.939   & 0.812   & 0.031
    & 0.817   & 0.875   & 0.736   & 0.073
    & 0.810   & 0.884   & 0.670   & 0.037\\
    \hline
    \#8 & & $\checkmark$ & & $\checkmark$ & $\{ 1,0,0 \}$ & $\{ 1;1;1 \}$
    & \Rev{\textbf{0.888}}   & 0.940   & 0.812   & 0.031
    & 0.819   & 0.877   & 0.741   & 0.072
    & 0.814   & 0.887   & 0.681   & 0.037\\
    \#9 & & $\checkmark$ & & $\checkmark$ & $\{ 1,0,0 \}$ & $\{ 8;8;8 \}$
    & 0.886   & 0.943   & 0.814   & 0.032
    & 0.816   & 0.872   & 0.738   & 0.074
    & \Rev{\textbf{0.815}}   & 0.886   & \Rev{\textbf{0.682}}  & 0.037\\
    \#10 & & $\checkmark$ & & $\checkmark$ & $\{ 1,0,0 \}$ & $\{ 32;32;32 \}$
    & 0.884   & \Rev{\textbf{0.944}}   & 0.810   & 0.033
    & 0.819   & 0.876   & 0.738   & 0.071
    & 0.813   & 0.884   & 0.675   & 0.037\\
    \#11 & & $\checkmark$ & & $\checkmark$ & $\{ 1,0,0 \}$ & $\{ 1;8;32 \}$
    & 0.883   & 0.940   & 0.812   & 0.032
    & 0.811   & 0.869   & 0.734   & 0.073
    & \Rev{\textbf{0.815}}   & 0.887   & 0.679   & \Rev{\textbf{0.036}}\\
    \rowcolor{mygray}
    \#OUR & & $\checkmark$ & & $\checkmark$ & $\{ 1,0,0 \}$ & $\{ 32;8;1 \}$
    & \Rev{\textbf{0.888}} & 0.942 & 0.816 & \Rev{\textbf{0.030}}
    & \Rev{\textbf{0.820}} & \Rev{\textbf{0.882}} & \Rev{\textbf{0.743}} & \Rev{\textbf{0.070}}
    & \Rev{\textbf{0.815}} & 0.887 & 0.680 & 0.037\\
    \bottomrule
    \end{tabular}
\end{table*}

\tabref{tab:DatasetGeneralization} provides the S-measure results 
for the cross-dataset generalization.
Each row lists a model that is trained on one dataset and tested on all others, 
indicating the generalizability of the dataset used for training.
Each column shows the performance of one model tested on a specific dataset and 
trained on all others,
indicating the difficulty of the testing dataset.
Please note that the training/testing settings are different from those used 
in \tabref{tab:ModelScore}, and thus the performances are not comparable.
\Rev{As expected, we find that our \ourdataset~has better generalization ability than the CAMO 
(\eg, the last column `\emph{Drop$\downarrow$:} -6.0\%')}.
This is because our dataset contains a variety of challenging concealed objects 
(\secref{sec:CODdataset}).
We can thus see that our COD10K dataset \Rev{contains more challenging scenes.}

\subsection{Ablation Studies}\label{sec:ablation_study}

We now provide a detailed analysis of the proposed \ournewmodel~on 
CHAMELEON, CAMO, and COD10K. We verify the effectiveness 
by decoupling various sub-components, including the NCD, TEM, and GRA, 
as summarized in~\tabref{tab:ablation}.
Note that we maintain the same hyperparameters mentioned in
\secref{sec:implementation_details} during the re-training process 
for each ablation variant.

\myPara{Effectiveness of NCD.}
We explore the influence of the decoder in the search phase of our 
\ournewmodel. 
To verify its necessity, we retrain our network without the NCD (No.\#1) 
and find that, compared with \#OUR (last row in \tabref{tab:ablation}), 
the NCD is attributed to boosting the performance on CAMO, 
increasing the mean $E_\phi$ score from 0.869 to 0.882. 
Further, we replace the NCD with the partial decoder~\cite{wu2019cascaded} 
(\ie, PD of No.\#2) to test the performance of this scheme.
%
Comparing No.\#2 with \#OUR, our design can enhance the performance slightly, 
increasing it by 1.7\% in terms of $F^w_\beta$ on the CHAMELEON.

\begin{figure}[t!]
  \centering
  \includegraphics[width=\linewidth]{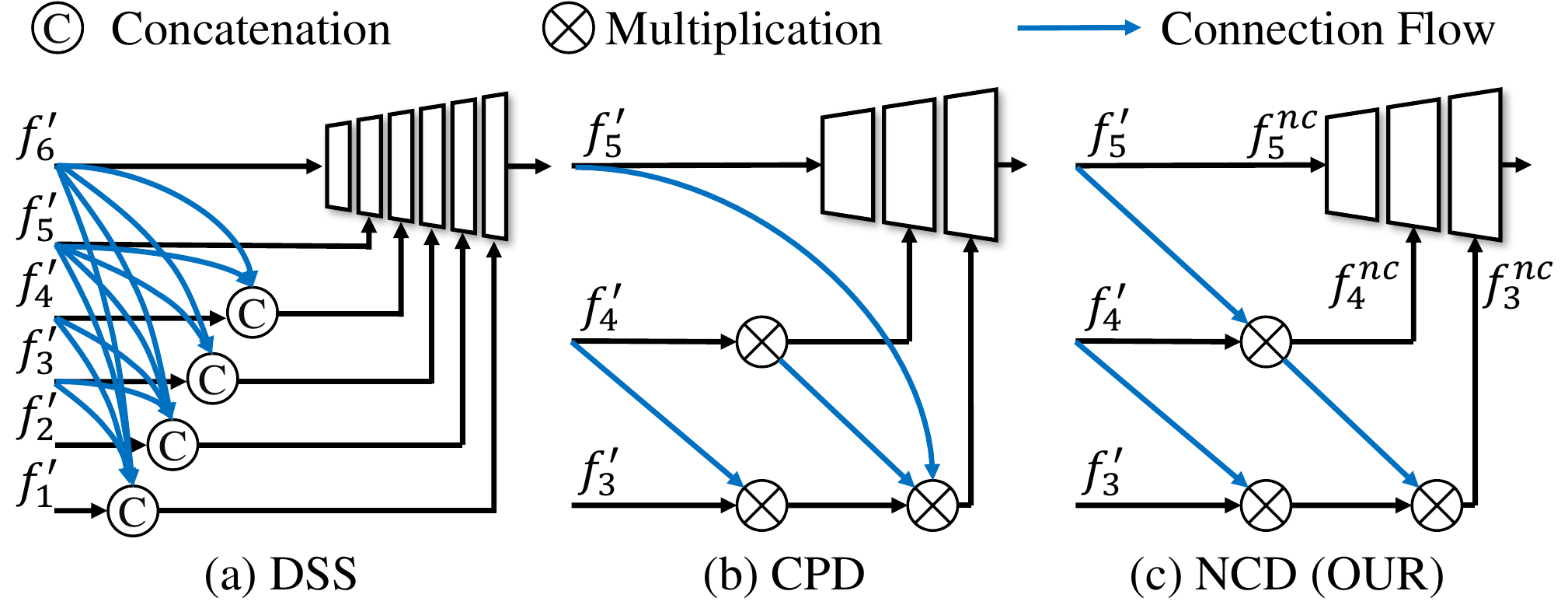}\\
  \vspace{-12pt}
  \caption{\textbf{Comparison of various types of inter-layer feature 
    aggregation strategies with a short connection.} 
    (a) DSS~\cite{HouPami19Dss} introduce the densely connected short 
    connection in a top-down manner. 
    (b) CPD~\cite{wu2019cascaded} constructs a partial decoder by discarding 
    larger resolution features of shallower layers for memory and speed 
    enhancement. 
    (c) Our neighbor connection decoder only propagates between neighboring
    layers. 
  }\label{fig:agg_strategy}
\end{figure}

As shown in~\figref{fig:agg_strategy}, we present a novel feature aggregation
strategy before the modified UNet-like decoder 
(removing the bottom-two high-resolution layers), termed the NCD, 
with neighbor connections between adjacent layers. 
This design is motivated by the fact that the high-level features are superior 
to semantic strength and location accuracy, 
but introduce noise and blurred edges for the target object.

Instead of broadcasting features from densely connected layers with a short 
connection~\cite{HouPami19Dss} or a partial decoder with a 
skip connection~\cite{wu2019cascaded}, 
our NCD exploits the semantic context through a \textit{neighbor connection},
providing a simple but effective way to reduce inconsistency between 
different features.
Aggregating all features by a \textit{short connection}~\cite{HouPami19Dss} 
increases the parameters. 
This is one of the major differences between DSS (\figref{fig:agg_strategy}~a) 
and NCD. 
Compared to CPD~\cite{wu2019cascaded} (\figref{fig:agg_strategy} b), 
which ignores feature transparency between $f_5'$ and $f_4'$, 
NCD is more efficient at broadcasting the features step by step.

\myPara{Effectiveness of TEM.}
We provide \Rev{two} different variants:
\textit{(a)} without TEM (No.\#3), and
\textit{(b)} with symmetric convolutional layers~\cite{szegedy2016rethinking}  
(No.\#4).
Comparing with No.\#3, we find that our TEM with asymmetric convolutional 
layers (No.\#OUR) is necessary for 
increasing the performance on the CAMO dataset.
Besides, replacing the standard symmetric convolutional layer (No.\#4) 
with an asymmetric convolutional layer (No.\#OUR) has little impact 
on the learning capability of the network, 
while further increasing the mean $E_\phi$ from 0.866 to 0.882 on the 
CAMO dataset.

\myPara{Effectiveness of GRA.}
\textit{Reverse Guidance.} 
As shown in the `Reverse' column of~\tabref{tab:ablation}, \{*,*,*\} 
indicates whether the guidance is reversed (see \figref{fig:GRA} (b)) 
before each GRA block $G^k_i$.
For instance, \{1,0,0\} means that we only reverse the guidance in the first 
block (\ie, $r^k_1$) and the remaining two blocks (\ie, $r^k_2$ and $r^k_3$) 
do not have a reverse operation.

We investigate the contribution of the reverse guidance in the GRA, 
including three alternatives:
\textit{(a)} without any reverse, \ie, \{0,0,0\} of No.\#5, 
\textit{(b)} reversing the first two guidances $r^k_i, i \in \{ 1, 2\}$, 
\ie, \{1,1,0\} of No.\#6, and 
\textit{(c)} reversing all the guidances $r^k_i, i \in \{ 1, 2, 3\}$, 
\ie, \{1,1,1\} of No.\#7. 
Compared to the default implementation of \ournewmodel~(\ie, \{1,0,0\} 
of No.\#OUR), 
we find that only reversing the first guidance may help the network 
to mine diversified representations from two perspectives 
(\ie, attention and reverse attention regions), 
while introducing reverse guidance several times in the intermediate process 
may cause confusion during the learning procedure, 
especially for setting \#6 on the CHAMELEON and COD10K datasets.

\textit{Group Size of GGO}.
As shown in the `Group Size' column of ~\tabref{tab:ablation}, 
$\{*;*;*\}$ indicates the number of feature slices (\ie, group size $g_i$) 
from the GGO of the first block $G^k_1$ to last block $G^k_3$.
For example, $\{ 32;8;1 \}$ indicates that we split the candidate feature 
$p^k_i, i \in \{ 1,2,3 \}$ into 32, 8, and 1 group sizes at each GRA block 
$G^k_i, i \in \{ 1,2,3 \}$, respectively. 
Here, we discuss two ways of selecting the group size, \ie, 
the uniform strategy (\ie, $\{1;1;1\}$ of \#8, $\{8;8;8\}$ of \#9, 
$\{32;32;32\}$ of \#10) and progressive strategy (\ie, $\{1;8;32\}$ of \#11 
and $\{32;8;1\}$ of \#OUR).
We observe that our design based on the progressive strategy can effectively 
maintain the generalizability of the network, 
providing more satisfactory performance compared with other variants.

\section{Downstream Applications}\label{sec:Applications}
Concealed object detection systems have various downstream applications 
in fields such as medicine, art, and agriculture. 
\Rev{Here, we envision some potential uses due to the common feature 
of these applications where the target objects share similar appearance 
with the background. 
Under such circumstances, COD models are very suitable to act as 
a core component of these applications to mine camouflaged objects.
Note that these applications are only toy examples to spark 
interesting ideas for future research.} 

\subsection{Application I: Medicine}

\subsubsection{Polyp Segmentation} 
\begin{figure}[t!]
  \centering
  \begin{overpic}[width=\columnwidth]{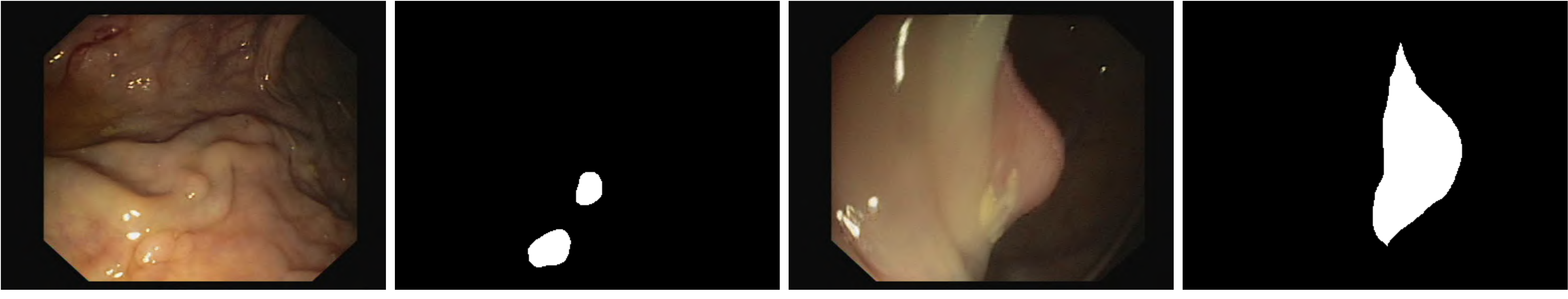}
	\put(10,-3){(a)}
	\put(35,-3){(b)}
	\put(62,-3){(c)}
	\put(85,-3){(d)}
  \end{overpic} 
  \vspace{-12pt}
  \caption{\textbf{Polyp segmentation.} 
    (a) \& (c) are input polyp images. 
    (b) \& (d) are corresponding ground-truths.
  }\label{fig:Polyp}
\end{figure}

As we all know, early diagnosis through medical imaging plays a key role in 
the treatment of diseases. 
However, the early disease area/lesions usually have a high degree of 
homogeneity with the surrounding tissues. 
As a result, it is difficult for doctors to identity the lesion area in the 
early stage from a medical image. 
One typical example is the early colonoscopy to segment polyps, 
which has contributed to roughly 30\% decline in the incidence of 
colorectal cancer~\cite{fan2020pranet}. Similar to concealed object detection, 
polyp segmentation (see \figref{fig:Polyp}) also faces several challenges, 
such as variation in appearance and blurred boundaries. 
The recent state-of-the-art polyp segmentation model, 
PraNet~\cite{fan2020pranet}, has shown promising performance in 
both \Rev{polyp} segmentation (Top-1) and concealed object segmentation (Top-2). 
\Rev{From this point of view, embedding our SINet into this application 
could potentially achieve more robust results.}

\begin{figure}[t!]
  \centering
  \begin{overpic}[width=\columnwidth]{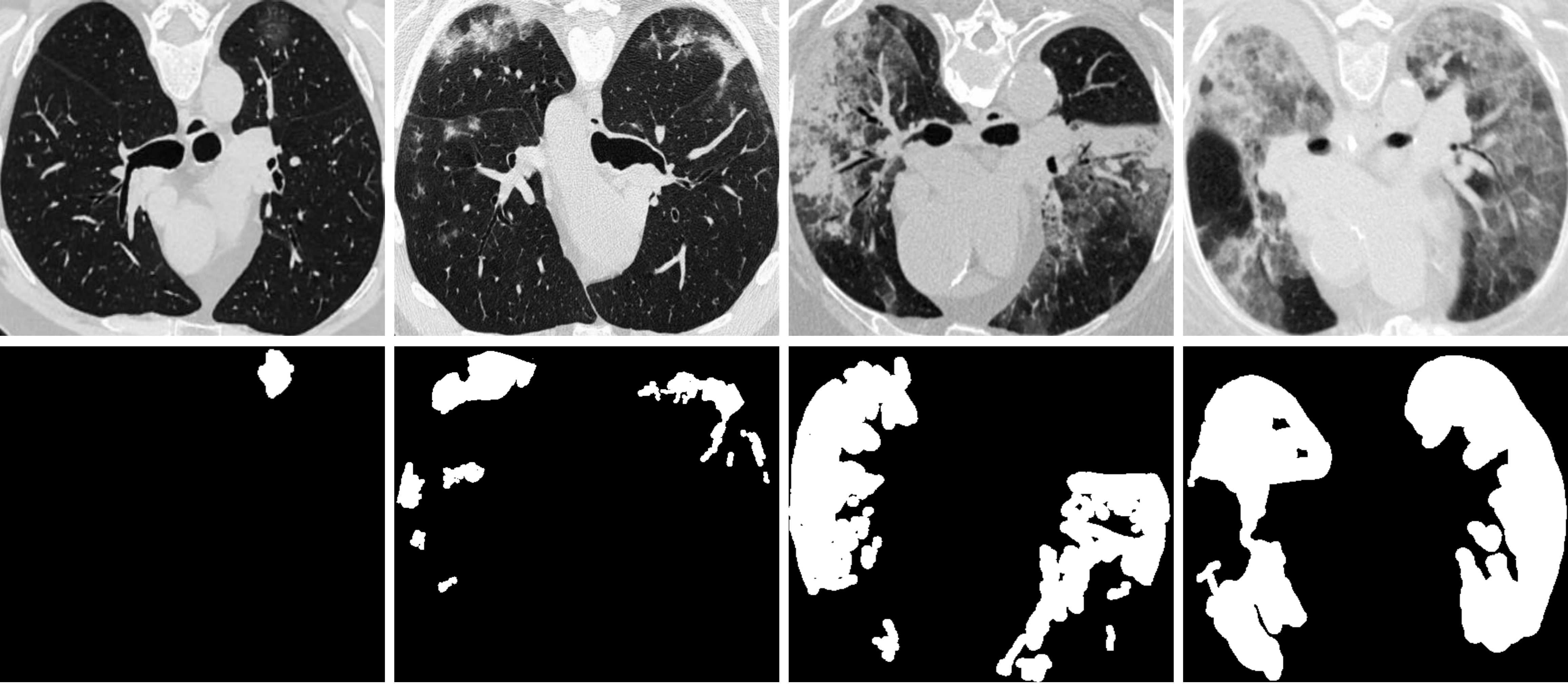}
	\put(10,-3){(a)}
	\put(35,-3){(b)}
	\put(62,-3){(c)}
	\put(85,-3){(d)}
  \end{overpic}
  \vspace{-12pt}
  \caption{\textbf{Lung infection segmentation.} 
    The first row presents COVID-19 lung infection CT scans, 
    while the second row shows their ground-truths labeled by doctors. 
    From (a) to (d), COVID-19 patients from mild to severe.  
  }\label{fig:Lung}
\end{figure}

\begin{figure}[t!]
  \centering
  \begin{overpic}[width=\columnwidth]{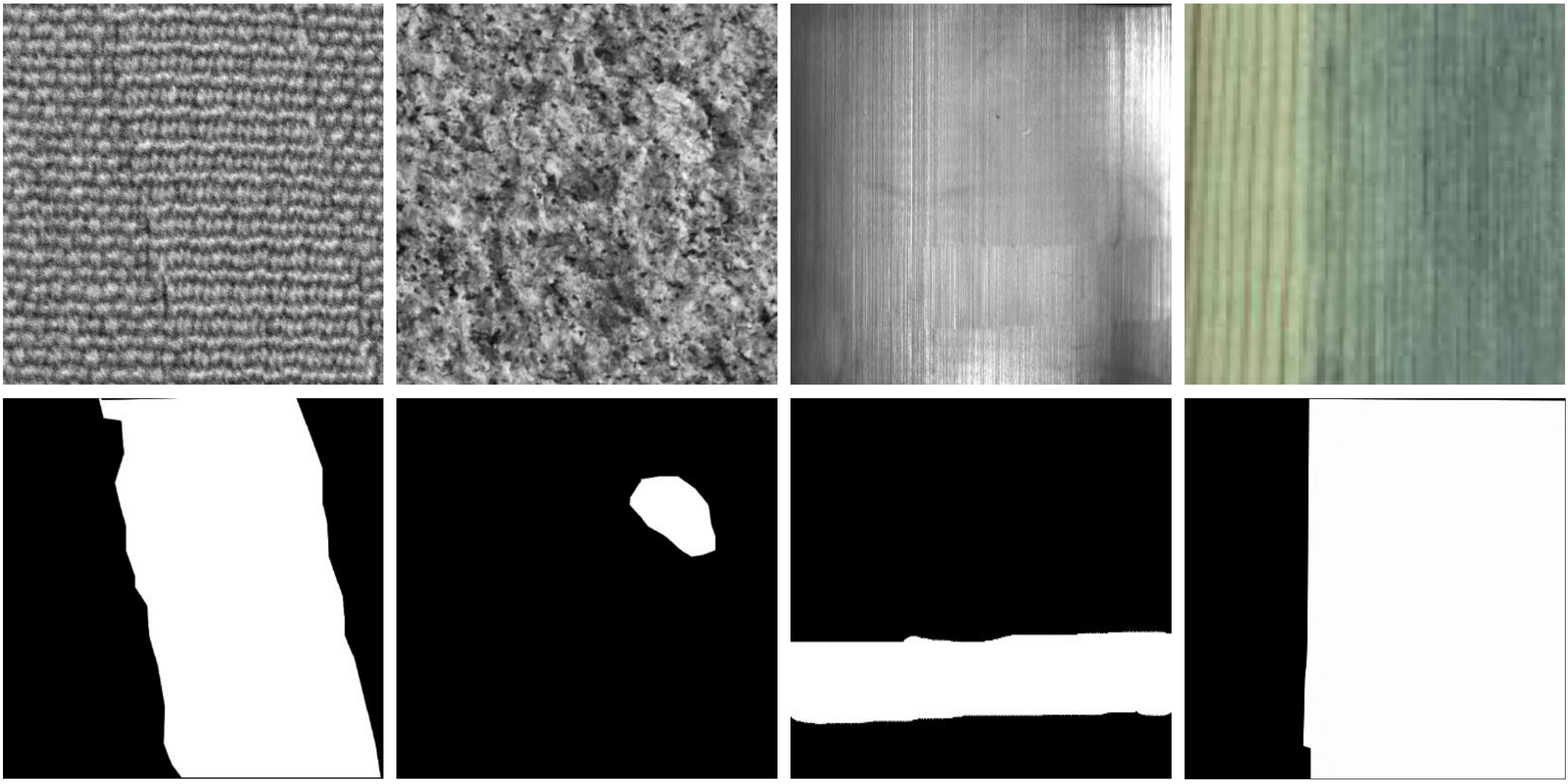}
   	\put(10,-3){(a)}
	\put(35,-3){(b)}
	\put(62,-3){(c)}
	\put(85,-3){(d)}
  \end{overpic}
  \vspace{-12pt}
  \caption{\textbf{Surface defect detection.}
	The defect types are textile (a), stone (b), magnetic tile (c), and wood (d), respectively. \Rev{The second row presented their corresponding ground truths.}
	Source images are derived from~\cite{he2019fully}.
  }\label{fig:SurfaceDefect}
\end{figure}
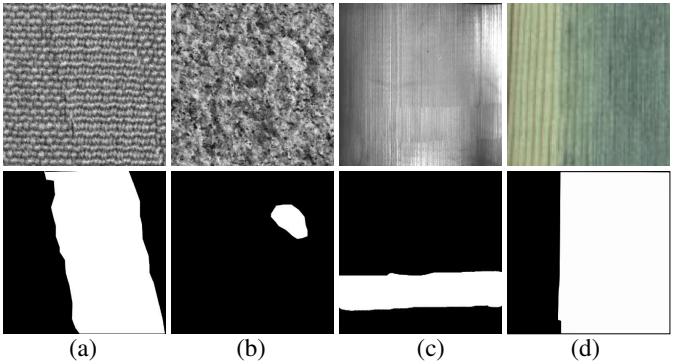

\subsubsection{Lung Infection Segmentation} 
Another concealed object detection example is the lung infection segmentation 
task in the medical field. Recently, 
COVID-19 has been of particular concern, and resulted in a global pandemic.
An AI system equipped with a COVID-19 lung infection segmentation model 
would be helpful in the early screening of COVID-19. 
More details on this application can be found in 
the recent segmentation model~\cite{fan2020inf} and survey paper~\cite{shi2020review}. \Rev{We believe retrain our SINet model using COVID-19 lung infection segmentation datasets will be another interesting potential application.}  

\subsection{Application II: Manufacturing}
\subsubsection{Surface Defect Detection}
In industrial manufacturing, products (\eg, wood, textile, and magnetic tile) 
of poor quality will inevitably lead to adverse effects on the economy. 
As can be seen from \figref{fig:SurfaceDefect}, 
the surface defects are challenging, 
with different factors including low contrast, ambiguous boundaries and so on.  
Since traditional surface defect detection systems mainly rely on humans, 
major issues are highly subjective and time-consuming to identify.  
Thus, designing an automatic recognition system based on AI is essential to 
increase productivity.
\Rev{We are actively constructing such a data set to advance related research.}
Some related papers can be found at: 
\url{https://github.com/Charmve/Surface-Defect-Detection/tree/master/Papers}.

\begin{figure}[t!]
  \centering
  \begin{overpic}[width=\columnwidth]{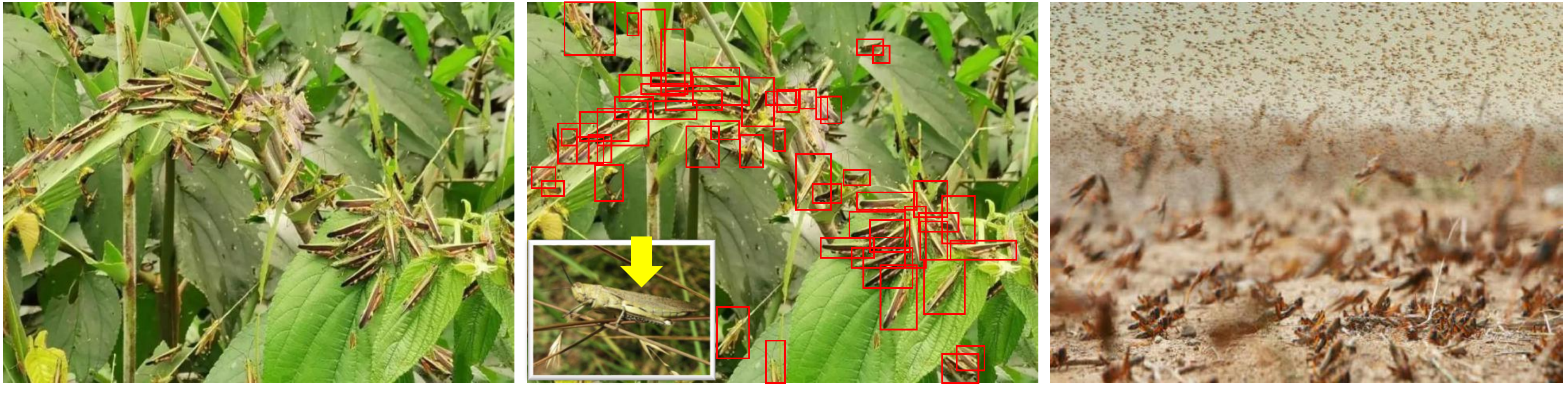}
    \put(14,-3){(a)}
	\put(48,-3){(b)}
	\put(81,-3){(c)}
  \end{overpic}
  \vspace{-12pt}
  \caption{\textbf{Pest detection.}
    For pest detection applications, the system can generate 
    a bounding box  (b) for each locally screened image (a) 
    or provide statistics (pest counting) for locust plague density monitoring 
    in the whole environment (c). 
  }\label{fig:PestsDetection}
\end{figure}

\begin{figure}[t!]
  \centering
  \begin{overpic}[width=\columnwidth]{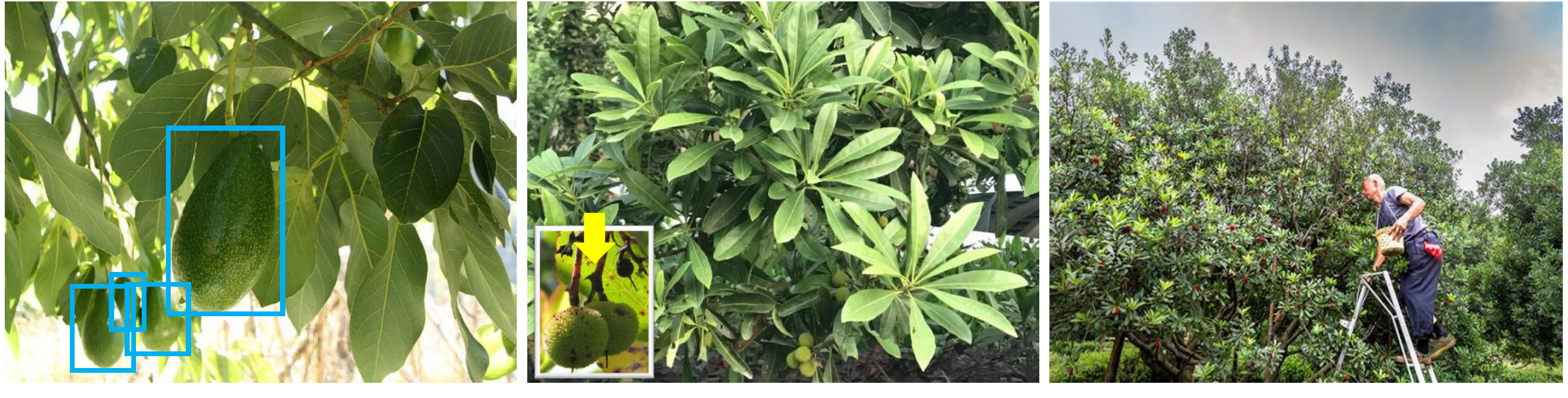}
    \put(14,-3){(a)}
	\put(48,-3){(b)}
	\put(81,-3){(c)}
  \end{overpic}
  \vspace{-12pt}
  \caption{\textbf{Fruit maturity detection.}
    Compared with the traditional manual inspection (c) of fruits, 
    such as Persea Americana (a) and Myrica Rubra (b) for maturity, 
    an AI-based maturity monitoring system will 
    greatly improve production efficiency. 
  }\label{fig:FruitMaturity}
\end{figure}

\subsection{Application III: Agriculture}
\subsubsection{Pest Detection}
Since early 2020, plagues of desert locusts have invaded the world, 
from Africa to South Asia. Large numbers of locusts gnaw on fields and 
completely destroy agricultural products, 
causing serious financial losses and famine due to food shortages. 
As shown in \figref{fig:PestsDetection}, 
introducing AI-based techniques to provide scientific monitoring is 
feasible for achieving sustainable regulation/containment by governments.
\Rev{Collecting relevant insect data for COD models requires rich biological knowledge, which is also a difficulty faced in this application.}

\subsubsection{Fruit Maturity Detection}
In the early stages of ripening, many fruits appear similar to green leaves,
making it difficult for farmers to monitor production. 
We present two types of fruits, \ie, Persea Americana and Myrica Rubra, 
in \figref{fig:FruitMaturity}. These fruits share similar characteristics 
to concealed objects, so it is possible to utilize a COD algorithm 
to identify them and improve the monitoring efficiency.

\begin{figure}[t!]
  \centering
  \includegraphics[width=\columnwidth]{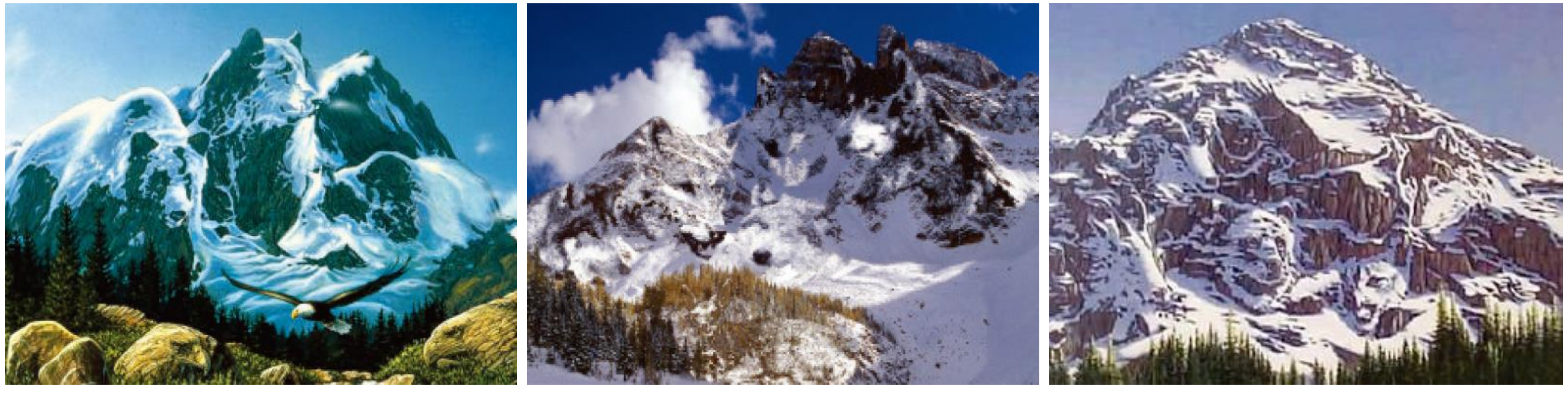} \\
  \vspace{-12pt}
  \caption{\textbf{Recreational art.}
	Some animals are embedded into the background by algorithms. 
    Source images from Chu \etal~\cite{chu2010camouflage} 
    and all rights reserved by 2010 John Van Straalen. 
  }\label{fig:Recreational}
\end{figure}

\subsection{Application IV: Art}
\subsubsection{Recreational Art}
Background warping to concealed salient objects is a fascinating technique 
in the SIGGRAPH community. 
\figref{fig:Recreational} presents some examples generated by Chu~\etal 
in~\cite{chu2010camouflage}. We argue that this technique will provide more 
training data for existing data-hungry deep learning models, 
and thus it is of value to explore the \Rev{underlying} mechanism behind the 
\textit{feature search} and \textit{conjunction search} theory described 
by Treisman and Wolfe~\cite{treisman1988features,wolfe1994guided}.

\begin{figure}[t!]
  \centering
  \begin{overpic}[width=\columnwidth]{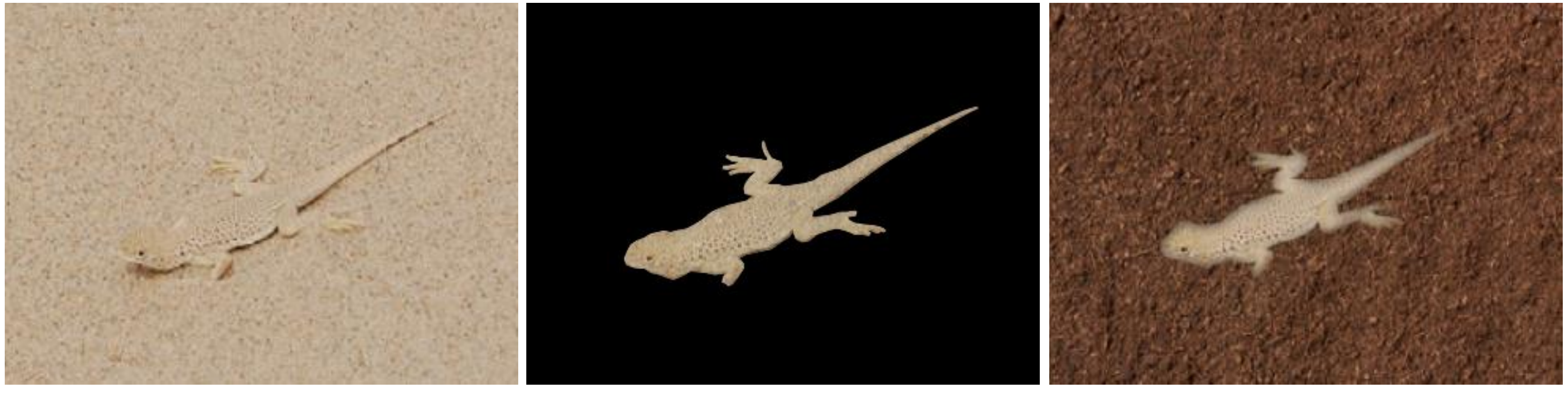}
    \put(14,-3){(a)}
	\put(48,-3){(b)}
	\put(81,-3){(c)}
  \end{overpic}
  \vspace{-12pt}
  \caption{\textbf{Converting concealed objects to salient objects.}
    Source images from~\cite{le2019anabranch}. 
    One interesting application is to identify (b) 
    a specific concealed object (a) and then convert it to a salient object (c).
  }\label{fig:Concealed2Salient}
\end{figure}

\begin{figure}[t!]
  \centering
  \begin{overpic}[width=\columnwidth]{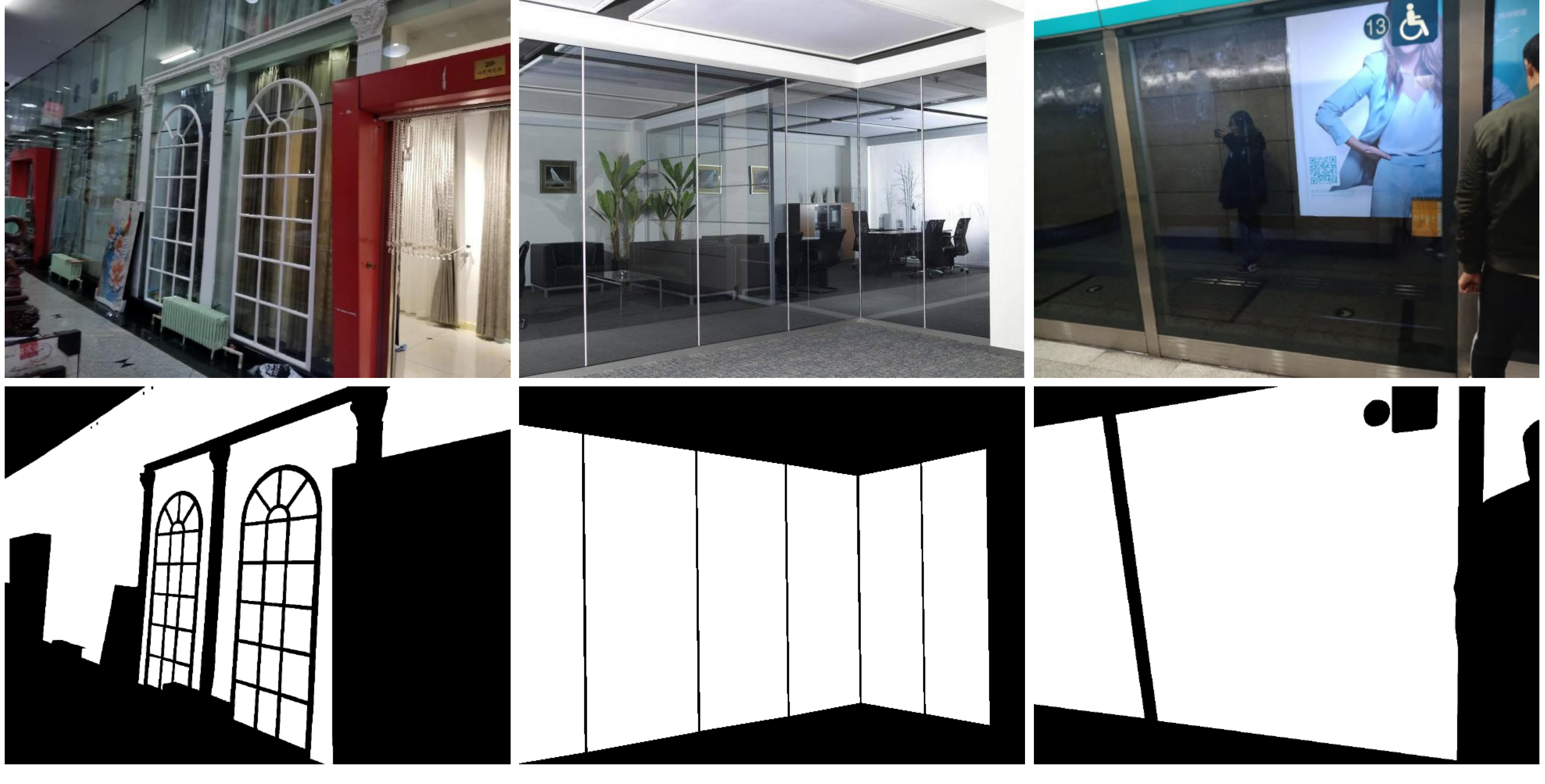}
    \put(14,-3){(a)}
	\put(48,-3){(b)}
	\put(81,-3){(c)}
  \end{overpic}
  \vspace{-12pt}
  \caption{\textbf{Transparent stuff/objects detection.}
    In our daily lives, we humans see, touch, or interact with various 
    transparent stuff such as windows (a), glass doors (b), 
    and glass walls (c). 
    \Rev{Second rows are corresponding ground-truths.}
    It is essential to teach AI robots to identify transparent stuff/objects 
    to avoid unseen obstacles.
  }\label{fig:TransparentObject}
\end{figure}

\begin{figure}[t!]
  \centering
  \begin{overpic}[width=\columnwidth]{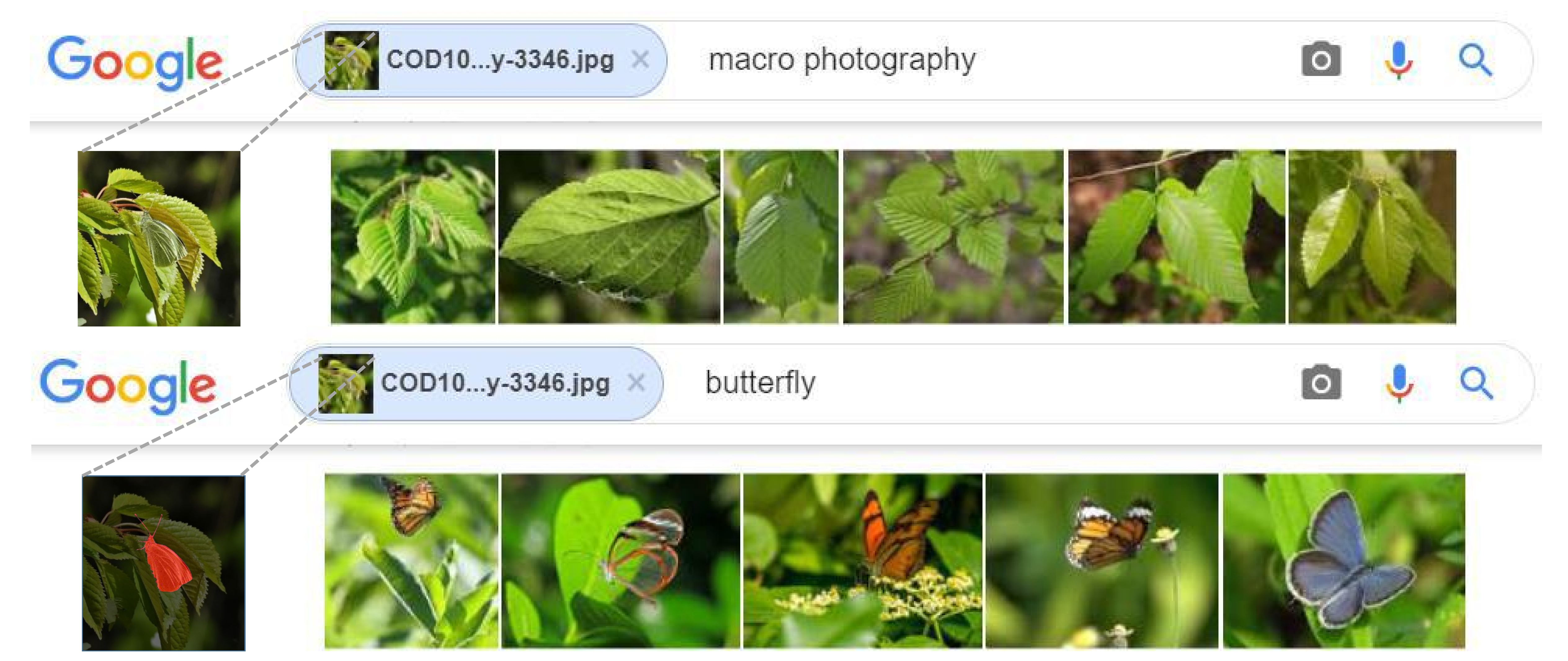}
    \put(0,5){(b)}
    \put(0,25){(a)}
  \end{overpic} \\
  \vspace{-12pt}
  \caption{\textbf{Search engines}. Internet search engine application 
    equipped without (a)/with (b) a concealed detection system.
  }\label{fig:SearchEngine}
\end{figure}

\subsubsection{From Concealed to Salient Objects}
Concealed object detection and salient object detection are two opposite tasks, 
making it convenient for us to design a multi-task learning framework that can 
simultaneously increase the robustness of the network. 
As shown in \figref{fig:Concealed2Salient}, 
there exist two reverse objects (a) and (c). 
An interesting application is to provide a scroll bar to allow users 
to customize the degree of salient objects from the concealed objects.

\subsection{Application V: Daily Life}
\subsubsection{Transparent Stuff/Objects Detection}

Transparent objects, such as glass products, are commonplace in our daily life. 
These \Rev{objects}/things, including doors and walls, 
\Rev{inherit} the appearance of their background, making them unnoticeable, 
as illustrated in \figref{fig:TransparentObject}. 
As a sub-task of concealed object detection, 
transparent object detection~\cite{xie2020segmenting} and 
transparent object tracking~\cite{fan2020transparent} have shown promise.

\subsubsection{Search Engines}
\figref{fig:SearchEngine} shows an example of search results from Google.
From the results (\figref{fig:SearchEngine} a), 
we notice that the search engine cannot detect
the concealed butterfly, and thus only provides images with similar backgrounds.
Interestingly, when the search engine is equipped with a concealed 
detection system (here, we just simply change the keyword),
it can identify the concealed object and then feedback several butterfly images 
(\figref{fig:SearchEngine} b).


\section{Potential Research Directions}\label{sec:Future}

Despite the recent 10 years of progress in the field of concealed object 
detection, 
the leading algorithms in the deep learning era remain limited compared to 
those for generic object detection~\cite{liu2020deep} and cannot yet 
effectively solve real-world challenges as shown in our \ourdataset~benchmark 
(Top-1: $F_\beta^w<0.7$). 
We highlight some long-standing challenges, as follows:
\begin{itemize}
\item Concealed object detection under limited conditions: 
few/zero-shot learning, weakly supervised learning, unsupervised learning, 
self-supervised learning, limited training data, unseen object class, \etc.

\item Concealed object detection combined with other modalities: 
Text, Audio, Video, RGB-D, RGB-T, 3D, \etc.

\item New directions based on the rich annotations provided in the \ourdataset,
such as concealed instance segmentation, concealed edge detection, 
concealed object proposal, concealed object ranking, among others.

\end{itemize}

Based on the above-mentioned challenges, there are a number of 
foreseeable directions for future research: 

\textbf{(1) Weakly/Semi-Supervised Detection:} 
Existing deep-based methods extract the features in a fully supervised manner 
from images annotated with object-level labels. 
However, the pixel-level annotations are usually manually marked by LabelMe 
or Adobe Photoshop tools with intensive professional interaction. 
Thus, it is essential to utilize weakly/semi (partially) annotated data 
for training in order to avoid heavy annotation costs. 

\textbf{(2) Self-Supervised Detection:} 
Recent efforts to learn representations (\eg, image, audio, and video)  
using self-supervised learning~\cite{afouras2020self,he2020momentum} 
have achieved  world-renowned achievements, attracting much attention. 
Thus, it is natural to setup a self-supervised learning benchmark for the 
concealed object detection task. 

\textbf{(3) Concealed Object Detection in Other Modalities:} 
Existing concealed data is only based on static images or dynamic videos
\cite{lamdouar2020betrayed}. 
However,  concealed object detection in other modalities can  be closely 
related in domains such as pest monitoring in the dark night, robotics, 
and artist design. 
Similar to in RGB-D SOD~\cite{Fan2019D3Net}, 
RGB-T SOD~\cite{zhang2019rgb}, CoSOD~\cite{deng2021re,Fan2021Group}, and VSOD~\cite{fan2019shifting}, 
these modalities can be audio, thermal, group image, or depth data, 
raising new challenges under specific scenes.  

\textbf{(4) Concealed Object Classification:} 
Generic object  classification is a fundamental task in computer vision. 
Thus concealed object classification will also likely gain attention 
in the future. 
By utilizing the class and sub-class labels provided in \ourdataset, 
one could build a large scale and fine-grain classification task. 

\textbf{(5) Concealed Object Proposal and Tracking:} 
In this paper, the concealed object detection is actually a segmentation task. 
It is different from traditional object detection, 
which generates a proposal or bounding boxes as the prediction. 
As such, concealed object proposal and tracking is a new and 
interesting direction~\cite{mondal2020camouflaged} for future work.

\textbf{(6) Concealed Object Ranking:} 
Currently, most concealed object detection algorithms are built upon binary ground-truths to generate the masks of concealed objects, 
\Rev{with only limited works analyzing the rank of concealed objects~\cite{zhai2021Mutual}.} 
However, understanding the level of concealment could help to better 
explore the mechanism behind the models, providing deeper insights into them. We refer readers to~\cite{kalash2019relative,zhai2021Mutual} for some inspiring ideas.

\textbf{(7) Concealed Instance Segmentation:} 
As described in~\cite{li2017instance}, 
instance segmentation is more crucial than object-level segmentation 
for practical applications.
\Rev{For example, we can push} the research 
on camouflaged object segmentation into camouflaged instance segmentation. 

\textbf{(8) Universal Network for Multiple Tasks:}
As studied by Zamir \etal in Taskonomy~\cite{zamir2018taskonomy}, 
different visual tasks have strong relationships. 
Thus, their supervision can be reused in one universal system without 
piling up complexity. It is natural to consider devising a universal network 
to simultaneously localize, segment and rank concealed objects. 
 
\textbf{(9) Neural Architecture Search:} 
Both traditional algorithms and deep learning-based models for concealed 
object detection require human experts with strong prior knowledge 
or skilled expertise. 
Sometimes, the hand-crafted features and architectures designed by 
algorithm engineers may not optimal. 
Therefore, neural architecture search techniques, 
such as the popular automated machine learning~\cite{yao2018taking}, 
offer a potential direction.

\textbf{(10) Transferring Salient Objects to Concealed Objects:}
Due to space limitations, we only evaluated typical salient object detection 
models in our benchmark section. 
There are several valuable problems that deserve further studying, 
however, such as transferring salient objects to concealed objects 
to increase the training data, and introducing a generative adversarial 
mechanism between the SOD and COD tasks to 
increase the feature extraction ability of the network.

The ten new research directions listed for concealed object remain far 
from being solved. 
However, there are many famous works that can be referred to, 
providing us a solid basis for studying the object detection task from 
a concealed perspective. 

\section{Conclusion}\label{sec:Conclusion}

We have presented the first comprehensive study on object detection from 
a concealed vision perspective.
Specifically, we have provided the new challenging and densely annotated
\ourdataset~dataset, conducted a large-scale benchmark,
developed a simple but efficient end-to-end search and identification 
framework (\ie, \ournewmodel),
and highlighted several potential applications.
Compared with existing cutting-edge baselines, our \ournewmodel~is
competitive and generates more visually favorable results.
%
The above contributions offer the community an opportunity to design 
new models for the COD task.
In the future, we plan to extend our \ourdataset~dataset to provide 
inputs of various forms, such as multi-view images 
(\eg, RGB-D SOD~\cite{fu2021siamese,zhang2021uncertainty}), 
textual descriptions, video (\eg, VSOD~\cite{fan2019shifting}), among others.
We also plan to automatically search the optimal receptive fields 
\cite{gao2021global2local} and employ improved feature representations 
\cite{gao2021rbn} for better model performance.

\begin{figure}[t!]
  \centering
  \includegraphics[width=\linewidth]{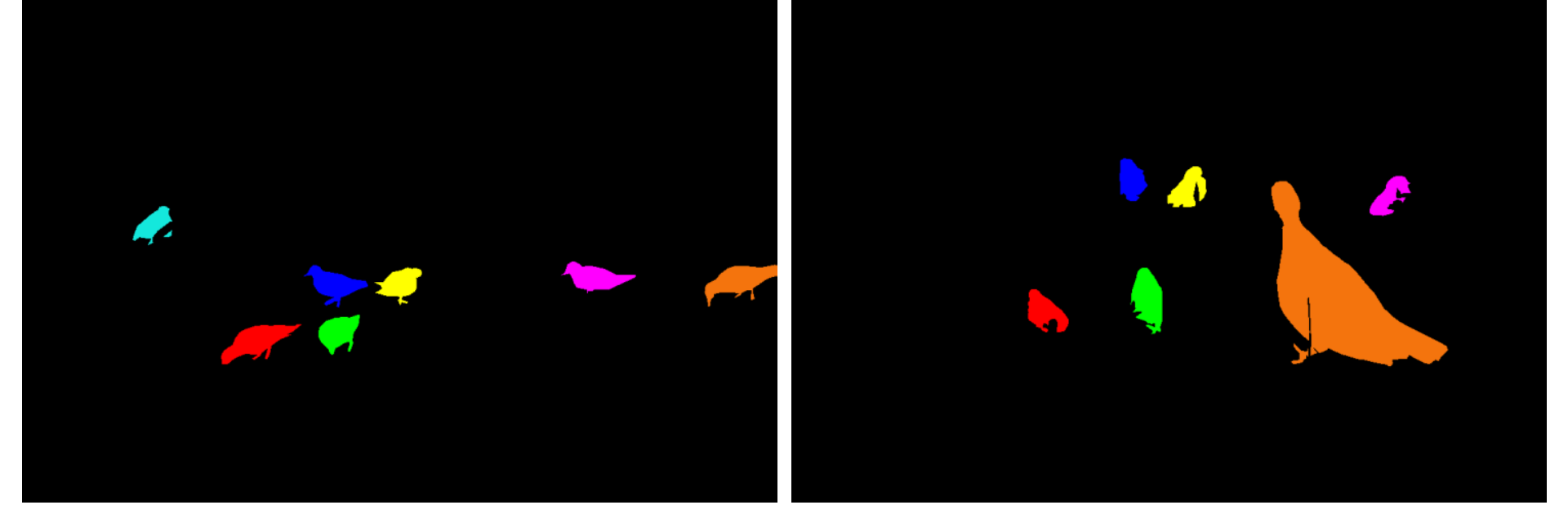}\\
  \vspace{-12pt}
  \caption{\textbf{Ground-truths} of the images presented in 
    \figref{fig:COD10KExample}
  }\label{fig:Answer}
\end{figure}

\section*{Acknowledgments}
We thank Guolei Sun and Jianbing Shen for insightful feedback.
This research was supported by the 
National Key Research and Development Program of China
under Grant No. 2018AAA0100400, NSFC (61922046), 
and S\&T innovation project from Chinese Ministry of Education.

\ifCLASSOPTIONcaptionsoff
  \newpage
\fi

{
\bibliographystyle{IEEEtran}
\bibliography{Camouflage}
}

\vspace{-.5in}
\begin{IEEEbiography}[{\includegraphics[width=1in,keepaspectratio]{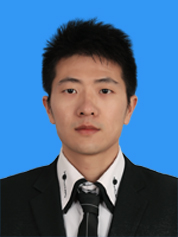}}]
{Deng-Ping Fan} received his PhD degree from the Nankai University in 2019.
He joined Inception Institute of AI in 2019.
He has published about 25 top journal and conference papers such as TPAMI, CVPR, ICCV, ECCV, \etc. 
His research interests include computer vision and visual attention, especially on 
RGB salient object detection (SOD), RGB-D SOD, Video SOD, Co-SOD. 
He won the Best Paper Finalist Award at IEEE CVPR 2019, 
the Best Paper Award Nominee at IEEE CVPR 2020.
\end{IEEEbiography}
\vspace{-.5in}

\begin{IEEEbiography}[{\includegraphics[width=1in,keepaspectratio]{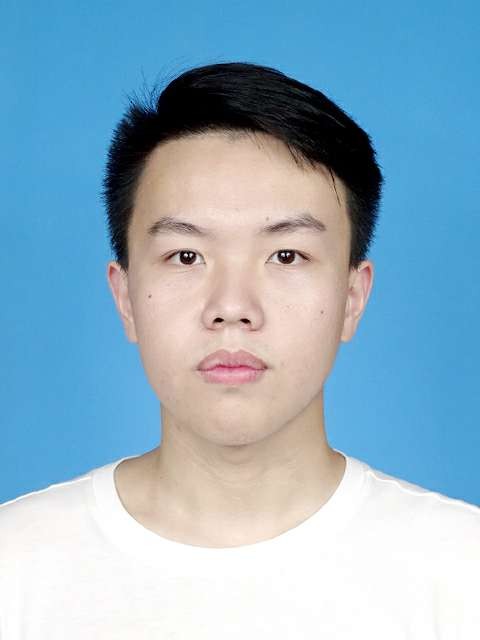}}]
{Ge-Peng Ji} is currently a MS of Communication and Information System at School of Computer Science, Wuhan University. His research interests lie in designing deep neural networks and applying deep learning in various fields of low-level vision, such as RGB salient object detection, RGB-D salient object detection, video object segmentation, concealed object detection, and medical image segmentation.
\end{IEEEbiography}
\vspace{-.5in}

\begin{IEEEbiography}[{\includegraphics[width=1in,keepaspectratio]{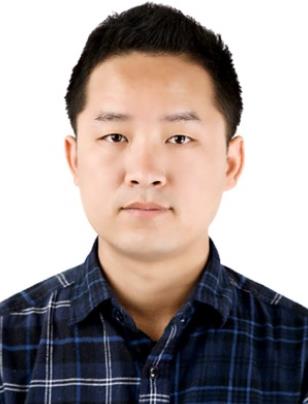}}]
{Ming-Ming Cheng} received his PhD degree from Tsinghua University in 2012. 
He then he did 2 years research fellow, with Prof. Philip Torr in Oxford. 
He is a full professor at Nankai University since 2016,
leading the Media Computing Lab. 
His research interests includes computer graphics, machine learning, 
computer vision, and image processing. 
He is an Associate Editor of IEEE TIP. 
He received several research awards, including the ACM China Rising Star Award, 
the IBM Global SUR Award, \etc.
\end{IEEEbiography}
\vspace{-.5in}

\begin{IEEEbiography}[{\includegraphics[width=1in,keepaspectratio]{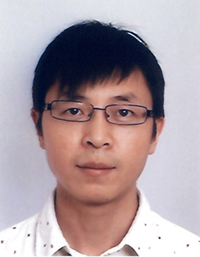}}]
{Ling Shao} is the CEO and the Chief Scientist of the Inception Institute of Artificial Intelligence (IIAI), Abu Dhabi, United Arab Emirates. He was the initiator and the Founding Provost and Executive Vice President of the Mohamed bin Zayed University of Artificial Intelligence (the world's first AI University), UAE. His research interests include computer vision, machine learning, and medical imaging. He is a fellow of the IEEE, the IAPR, the IET, and the BCS. 
\end{IEEEbiography}



\end{document}